\documentclass{article}

\PassOptionsToPackage{numbers, compress}{natbib}

\usepackage[final]{tackling_climate_workshop_style}

\usepackage[utf8]{inputenc} 
\usepackage[T1]{fontenc}    
\usepackage{hyperref}       
\usepackage{url}            
\usepackage{booktabs}       
\usepackage{amsfonts}       
\usepackage{nicefrac}       
\usepackage{microtype}      

\usepackage{multirow}
\usepackage{enumitem}
\usepackage{amsmath}
\usepackage{graphicx}
\usepackage{svg}

\usepackage{algorithm}
\usepackage{algpseudocode}

\usepackage{subcaption}
\usepackage{booktabs}

\title{RAIN: Reinforcement Algorithms for Improving Numerical Weather and Climate Models}

\author{%
Pritthijit Nath$^{1}$ \quad Henry Moss$^{1}$ \quad Emily Shuckburgh$^{2}$ \quad Mark Webb$^{3}$\\ 
$^1$ Department of Applied Math and Theoretical Physics, University of Cambridge\\
$^2$ Department of Computer Science and Technology, University of Cambridge\\
$^3$ Met Office Hadley Centre\\
\texttt{\{pn341,hm493,efs20\}@cam.ac.uk}; \texttt{mark.webb@metoffice.gov.uk}\\
}

\newif\ifshowcontent
\newif\ifnotshowcontent

\showcontenttrue 
\notshowcontentfalse

\begin{document}
\setcitestyle{square}

\maketitle

\begin{abstract}
This study explores integrating reinforcement learning (RL) with idealised climate models to address key parameterisation challenges in climate science. Current climate models rely on complex mathematical parameterisations to represent sub-grid scale processes, which can introduce substantial uncertainties. RL offers capabilities to enhance these parameterisation schemes, including direct interaction, handling sparse or delayed feedback, continuous online learning, and long-term optimisation. We evaluate the performance of eight RL algorithms on two idealised environments: one for temperature bias correction, another for radiative-convective equilibrium (RCE) imitating real-world computational constraints. Results show different RL approaches excel in different climate scenarios with exploration algorithms performing better in bias correction, while exploitation algorithms proving more effective for RCE. These findings support the potential of RL-based parameterisation schemes to be integrated into global climate models, improving accuracy and efficiency in capturing complex climate dynamics. Overall, this work represents an important first step towards leveraging RL to enhance climate model accuracy, critical for improving climate understanding and predictions. Code accessible at %
\ifshowcontent{
\url{https://github.com/p3jitnath/climate-rl}.}\fi
\ifnotshowcontent{\url{https://anonymous.4open.science/r/climate-rl-1F44}.}\fi
\end{abstract}

\section{Introduction}

Weather and climate modelling are crucial for understanding and mitigating the socio-economic impacts of meteorological phenomena. The UK has experienced an increase in extreme weather events, with the 10 warmest years on record occurring since 2003~\cite{met_office_2023_2024}. These events have caused significant disruptions, exemplified by the 2019-20 UK floods~\cite{sefton_20192020_2021} resulting in economic losses of at least £333 million~\cite{uk_government_environment_agency_national_2022}. Accurate forecasting and climate projections are essential for various sectors, underscoring their importance in national resilience and economic stability. Traditionally, Numerical Weather Prediction (NWP) methods have been utilised in weather and climate models, solving equations governing atmospheric dynamics to generate forecasts. Notable examples include the Met Office's Unified Model~\cite{met_office_unified_1990} and the ECMWF's Integrated Forecast System~\cite{ecmwf_integrated_2023}. While NWP models have improved forecast accuracy since their introduction in 1952~\cite{met_office_history_2017}, they still face limitations in representing sub-grid scale processes and resolving intricate atmospheric phenomena~\cite{lupo_global_2013}.

Although recent advancements in machine learning (ML) techniques have demonstrated strong potential in outperforming NWP models in short to medium-range weather forecasting~\cite{lang_aifs_2024, vaughan_aardvark_2024, stock_diffobs_2024, nguyen_climax_2023, lessig_atmorep_2023, ben-bouallegue_rise_2023, lam_learning_2023, price_gencast_2023, bi_accurate_2023, chen_fengwu_2023, nath_forecasting_2023, pathak_fourcastnet_2022, keisler_forecasting_2022}, purely data-driven artificial intelligence (AI) approaches face inherent limitations in abiding by fundamental laws of physics and thermodynamic principles~\cite{willard_integrating_2022, karpatne_theory-guided_2017}, and in generalising to out-of-distribution scenarios~\cite{lazer_parable_2014, caldwell_statistical_2014}. ML models also can produce outputs that violate key constraints like conservation laws of mass and energy~\cite{hansen_learning_2024, readshaw_incorporation_2023}, often requiring additional constraints to circumvent such violations~\cite{karniadakis_physics-informed_2021, beucler_enforcing_2021, bolton_applications_2019}. In long range climate projections, these violations can trigger instabilities and escalating errors, leading to rapid degradation of forecast skill beyond shorter lead times~\cite{de_burgh-day_machine_2023}.

To address these issues, our research explores the implementation of a reinforcement learning (RL) assisted climate modelling framework. RL algorithms learn by interacting with an environment, iteratively adjusting their behaviour to maximise a cumulative future reward signal. By formulating the climate model parameterisation problem as a control task, RL agents can learn to dynamically adjust the parameters of low resolution NWP models as a function of the atmospheric state, respecting the physics by leaving the structure of the physical parametrisations intact. For this research, we adopt a step-wise approach, first using RL for climate bias correction in a simple heating environment, then extending our methods to a climlab~\cite{rose_climlab_2018} based setup of radiative-convective equilibrium (RCE). In particular, this research advances the field of AI-based atmospheric modelling through the following key contributions:

\begin{enumerate}
    \item \textbf{Novel application of RL in climate modelling}: We explore the use of continuous action model-free RL algorithms  (REINFORCE~\cite{williams_simple_1992}, DDPG~\cite{lillicrap_continuous_2019}, DPG~\cite{silver_deterministic_2014}, TD3~\cite{fujimoto_addressing_2018}, PPO~\cite{schulman_proximal_2017}, TRPO~\cite{schulman_trust_2017}, SAC~\cite{haarnoja_soft_2018}, TQC~\cite{kuznetsov_controlling_2020}) to dynamically adjust parameters in climate models, offering a new approach to the long-standing challenge of parameterisation in NWP systems.
    \item \textbf{Integration of physical constraints}: Our framework demonstrates how RL can be employed to optimise model performance while adhering to fundamental physical constraints, addressing a critical limitation of purely data-driven approaches.
    
\end{enumerate}

\section{Methodology}

\subsection{Background}

\subsubsection{Advantages over traditional ML models}

RL offers three key advantages that are beneficial for improving  model parameterisations in the climate context:

\begin{enumerate}
\item \textbf{Continuous incremental learning}: RL algorithms can continuously adapt and update their policies through direct interaction with the climate model environment, unlike traditional static ML techniques that rely on fixed training datasets. This allows the RL agent to continuously refine the parameterisation scheme incrementally as the climate model evolves without the need for full retraining of the entire model, making the the process more computationally efficient and responsive to changes.

\item \textbf{Learning sparse rewards}: RL excels at learning from sparse or delayed rewards, which is particularly relevant for climate modelling, where reanlaysis data such as ERA5~\cite{hersbach_era5_2020} and satellite observations are available at six-hourly intervals. 

\item \textbf{Long-term optimisation}: RL focuses on maximising long-term rewards, which aligns well with the broader goal of climate modelling to understand and predict long-term climate patterns and trends. RL algorithms can effectively balance exploration (attempting new strategies) and exploitation (prioritising strategies with established outcomes) to find optimal parameterisation schemes.

\end{enumerate}

By leveraging these three main capabilities, RL can help climate researchers develop more dynamic and responsive parameterisations with the ability to continuously improve the accuracy and reliability of climate models in a computationally efficient manner over time. 

\subsubsection{Radiative-Convective Equilibrium}

Radiative-convective equilibrium (RCE) \cite{rose_12_2022, hartmann_chapter_2016, held_19_2011} is an idealised climate model balancing radiative (absorption and emission of radiation) and convective fluxes (vertical transport of energy) in a state of equilibrium. It solves for a single vertical temperature profile $T$ as a function of height $z$, representing the global average, assuming hydrostatic balance where the vertical pressure gradient ${\partial p}/{\partial z}$ is balanced by the gravitational force $g$ acting on the air density $\rho$.

\vspace{2cm}

The model comprises two main components:
\begin{enumerate}
    \item \textbf{Radiative transfer ($\mathcal{R}$)}: This calculates net shortwave and longwave fluxes based on incident solar radiation, surface albedo, and vertical profiles of temperature and radiatively active constituents. 
    \item \textbf{Convective adjustment ($\mathcal{D}$)}: This process maintains the lapse rate $-\partial T/\partial z$ below a critical value, empirically determined to be 6.5 K/km, by redistributing energy vertically while conserving mean temperature.
\end{enumerate}

\subsection{RL Environments}

\subsubsection{\texttt{SimpleClimateBiasCorrectionEnv}}
\label{section:scbc_env}

This environment (Fig.~\ref{fig:scbc_thermometer}) models temperature evolution, aiming to learn an optimal heating increment $u$ to minimise a bias correction term which relaxes the temperature to an observed value. Temperature dynamics (over 200 timesteps) are governed by:
\begin{align}
T_{\text{new}} = T_{\text{current}} + u +
\left(\frac{T_{\text{physics}} - T_{\text{current}}}{T_{\text{physics}} - T_{\text{observed}}}\right)\times 0.2 \\
T_{\text{new}} = T_{\text{new}} +
\left(\frac{T_{\text{observed}} - T_{\text{new}}}{T_{\text{physics}} - T_{\text{observed}}}\right)\times 0.1
\label{eq:scbc_new_temp}
\end{align}

The environment state is the current temperature, and the action space represents the bounded heating rate (\texttt{-1 - +1}). The reward signal in \texttt{v0} is:
\begin{equation}
\text{REWARD} = -1 \times \left[\left(\frac{T_{\text{observed}} - T_{\text{new}}}{T_{\text{physics}} - T_{\text{observed}}}\right)\times 0.1\right]^{2}
\label{eq:scbc_reward_v0}
\end{equation}

The environment supports graphical rendering using Pygame~\cite{mcgugan_introducing_2007}.
Versions \texttt{v1} and \texttt{v2} increase difficulty progressively. \texttt{v1} calculates reward as a mean squared error between $T_{\text{observed}}$ and $T_{\text{current}}$. \texttt{v2} introduces sparse rewards with a 5-timestep lag.

\subsubsection{\texttt{RadiativeConvectiveModelEnv}}
This environment (Fig.~\ref{fig:rcem_thermplots}) explores an idealised RCE model (implemented using climlab~\cite{rose_climlab_2018}) integrating over 500 timesteps. The action space comprises two parameters: emissivity (\texttt{0 - 1}) and adjusted lapse rate (\texttt{5.5 - 9.8}), while the state represents the air temperature profile across 17 pressure levels (1000 hPa - 10 hPa).
The model incorporates functions \texttt{climlab.radiation.RRTMG} and \texttt{climlab.convection.ConvectiveAdjustment} for the radiative transfer and convective adjustment processes respectively. After initialisation at isothermal equilibrium (constant temperature across all pressure levels), at each step, the environment updates parameters based on the RL agent's action, steps forward in time, and calculates cost (-ve reward) as the mean squared difference between simulated and observed temperature profiles (eg., NCEP/NCAR reanalysis~\cite{noaa_physical_sciences_laboratory_monthly_2011, kalnay_ncepncar_1996}). 

\section{Results}


\ifshowcontent{
\begin{table}[!h]
\centering
\caption{Frequency of RL algorithms within Top-3}
\label{tbl:results_topk}
\begin{subtable}{0.48\textwidth}
\centering
\caption{\texttt{SimpleClimateBiasCorrectionEnv}}
\label{tbl:results_scbc_topk}
\ttfamily
\begin{tabular}{ccc}
\toprule
\textbf{\textrm{Rank}} & \textbf{\textrm{Algorithm}} & \textbf{\textrm{Frequency}} \\ 
\midrule
1 & \textrm{DDPG} & 11 \\ 
2 & \textrm{TD3} & 10 \\ 
2 & \textrm{TQC} & 10 \\ 
4 & \textrm{DPG} & 3 \\ 
5 & \textrm{SAC} & 2 \\ 
\bottomrule
\end{tabular}
\end{subtable}
\begin{subtable}{0.48\textwidth}
\centering
\caption{\texttt{RadiativeConvectiveModelEnv}}
\label{tbl:results_rcem_topk}
\ttfamily
\begin{tabular}{ccc}
\toprule
\textbf{\textrm{Rank}} & \textbf{\textrm{Algorithm}} & \textbf{\textrm{Frequency}} \\ 
\midrule
1 & \textrm{DPG} & 4 \\ 
2 & \textrm{PPO} & 3 \\ 
3 & \textrm{TRPO} & 2 \\ 
3 & \textrm{TQC} & 2 \\
5 & \textrm{DDPG} & 1 \\
\bottomrule
\end{tabular}
\end{subtable}
\end{table}
}\fi

In all versions of \texttt{SimpleClimateBiasCorrectionEnv}, off-policy actor-critic methods consistently performed best, as shown in Table~\ref{tbl:results_scbc_topk}. The success of these algorithms, particularly DDPG, TD3, and TQC, can be attributed to their use of experience replay, which enhances sample efficiency and learning from diverse scenarios. Additionally, their actor-critic architecture, along with enhancements like target networks and delayed policy updates, also contributed to training stability. TQC's success, leveraging quantile regression, indicates the importance of risk-sensitive learning in environments with uncertainties common in climate simulations. Use of the quantile Huber loss function also improved robustness against outliers and training stability. 

For the \texttt{RadiativeConvectiveModelEnv}, on-policy methods dominated (Table~\ref{tbl:results_rcem_topk}). TRPO and PPO's success suggests that the RCE environment benefits from synchronised update strategies and gradual policy refinement. Their trust region approach, which restricts policy updates to minor adjustments, proved effective in an environment with sensitive dynamics or reward structures. This aligns with the RCE environment's design, featuring three distinct optima (Radiative-Convective, Radiative, and isothermal equilibrium), which favours strategies emphasising steady, incremental improvements over large, uncertain explorations. 

\begin{figure}[!h]
\centering
\begin{subfigure}{0.48\textwidth}
\centering
    \includegraphics[height=7cm]{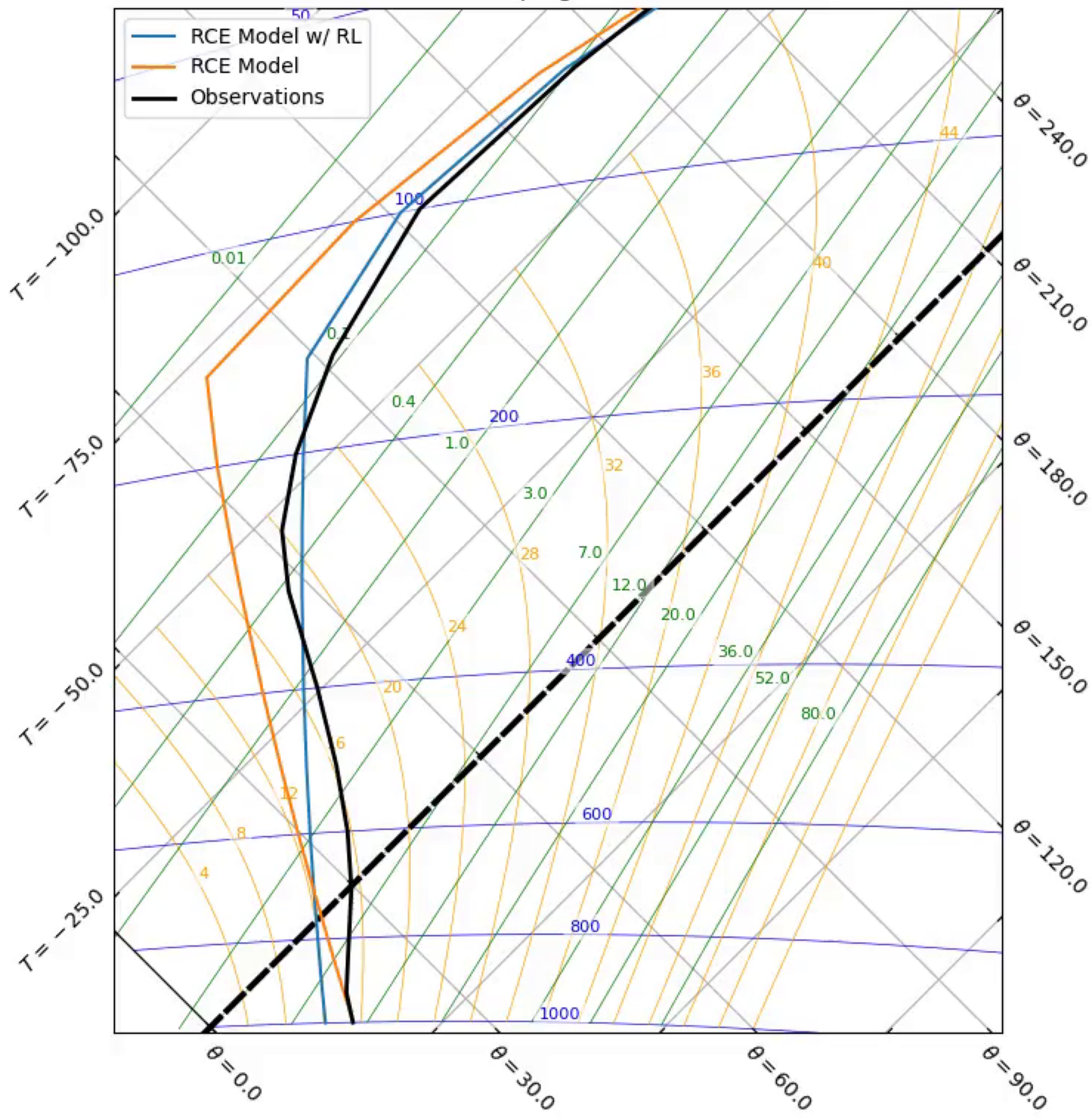}
\subcaption{Tephigram}
\end{subfigure}
\begin{subfigure}{0.48\textwidth}
\centering
    \includegraphics[height=7cm]{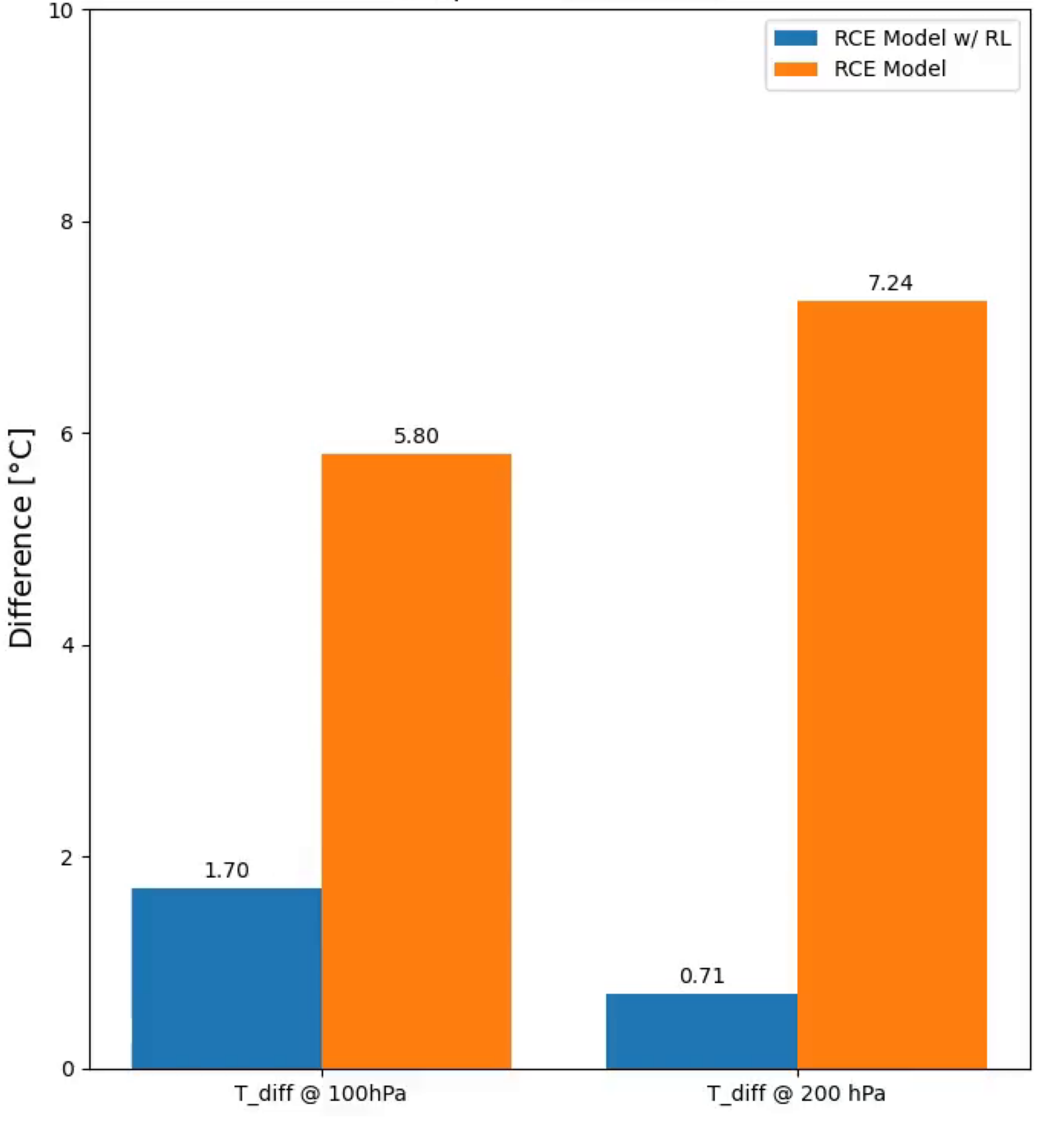}
\subcaption{Temperature differences}
\end{subfigure}
\caption{Snapshot of the temperature profile and the differences at 100 hPa and 200 hPa in the last episodic timestep of an RL-assisted RCE integration run using DPG in \texttt{rce-v0-optim-L-10k}.}
\label{fig:rcem_snapshot}
\end{figure}

The RL-assisted RCE Model demonstrates significant improvements in temperature difference at [\texttt{100 hPa, 200 hPa}], achieving [\texttt{70.87\%, 90.19\%}] improvement over 500 timesteps (Fig.~\ref{fig:rcem_snapshot}). This showcases superior tracking of the observation profile compared to the conventional RCE model, despite potential shortcomings at 1000 hPa.

The success of different RL algorithms in \texttt{SimpleClimateBiasCorrectionEnv} and \texttt{RadiativeConvectiveModelEnv} (Table ~\ref{tbl:results_scbc_topk} and ~\ref{tbl:results_rcem_topk}) highlights the importance of environment dynamics in algorithm selection. Off-policy algorithms like DDPG excel in scenarios with a diverse loss landscape (\texttt{SimpleClimateBiasCorrectionEnv}) where more exploration is required, while on-policy trust-region based algorithms like PPO perform better in environments with limited minima, where convergence can be initialisation sensitive (\texttt{RadiativeConvectiveModelEnv}). This insight is crucial for efficient algorithm selection in compute-constrained climate model simulations while using RL to enhance parameter adjustments and model performance in complex climate scenarios.

\enlargethispage{\baselineskip}

\section{Conclusion}

This study demonstrates RL's potential to enhance parameterisations in idealised climate models, representing one of the first efforts in computational climate science. Through extensive experimentation with continuous action model-free RL algorithms across two distinct idealised climate environments, we show that RL agents exhibit great promise in the dynamic setting of parameters (as a function of the atmospheric state) whilst maintaining the structure of the physical parametrisations intact. Off-policy algorithms (DDPG, TD3, TQC) excel in \texttt{SimpleClimateBiasCorrectionEnv}, while on-policy algorithms (DPG, PPO, TRPO) perform better in \texttt{RadiativeConvectiveModelEnv}. Our high-performance parallel computing setup (Section~\ref{sec:app-hyperparameter_tuning}) enables scalable RL experiments while meeting real-life climate model compute demands. This research also sets the stage for exploring RL in more complex scenarios, potentially integrating with the Met Office's weather and climate models. %
\ifshowcontent{
Though currently limited in scope, our work presents an initial step towards AI-assisted modelling that integrates RL with climate models, aiming to generate substantial research interest within the scientific ML community.
}\fi

\begin{ack}
P. Nath was supported by the \href{https://ai4er-cdt.esc.cam.ac.uk/}{UKRI Centre for Doctoral Training in Application of Artificial Intelligence to the study of Environmental Risks} [EP/S022961/1]. Mark Webb was supported by the Met Office Hadley Centre Climate Programme funded by DSIT. We thank Dr Sujit Roy (NASA-IMPACT) for highlighting an error in one of the equations in Section~\ref{section:scbc_env}.
\end{ack}

\bibliographystyle{vancouver}
\bibliography{bibliography_zotero}

\appendix
\renewcommand\thesection{Appendix \Alph{section}}
\renewcommand\thesubsection{\Alph{section}.\arabic{subsection}} 
\renewcommand\thetable{\Alph{section}.\arabic{table}}  
\renewcommand\thefigure{\Alph{section}.\arabic{figure}}  
\setcounter{table}{0}
\setcounter{figure}{0}

\clearpage

\section{Additional Background}

\subsection{RL Algorithm Summaries}

\begin{table}[!h]
\centering
\small
\caption{Four point summaries of key RL algorithms used}
\vspace{0.25em}
\label{tbl:rl_four_point_summaries}
\begin{tabular}{lp{11cm}}
\toprule
\textbf{Algorithm} & \textbf{Properties} \\ \midrule
REINFORCE~\cite{williams_simple_1992} & 1. Off-policy (requires a full trajectory) and non actor-critic. \\
& 2. Uses a stochastic policy to generate actions. \\
& 3. Uses policy-gradient for actor updates. \\
& 4. Samples trajectories and computes the discounted cumulative rewards (returns). \\ \midrule
Deterministic Policy & 1. On-policy and actor-critic. \\
Gradient (DPG)~\cite{silver_deterministic_2014} & 2. Uses deterministic policies (non $\epsilon$-greedy stochastic). \\
& 3. Uses temporal difference (TD) learning~\cite{sutton_reinforcement_2018} for critic and policy gradients~\cite{sutton_reinforcement_2018} for actor updates. \\
& 4. No target networks for actor or critics. \\ \midrule
Deep Deterministic & 1. Off-policy and actor-critic. \\
Policy Gradient (DDPG) & 2. Uses deterministic policies (non $\epsilon$-greedy stochastic). \\
\cite{lillicrap_continuous_2019} & 3. Uses an experience replay buffer to store past experiences. \\
& 4.~Uses target networks for both the actor and the critic to stabilise training. \\ \midrule
Twin Delayed DDPG & 1. Off-policy and actor-critic. \\
(TD3)~\cite{fujimoto_addressing_2018} & 2.~Uses two critic networks to reduce the overestimation bias in value estimation. \\
& 3. Updates the policy less frequently than the value function aiding stabilisation. \\
& 4. Adds gaussian noise to the target action to facilitate exploration. \\ \midrule
Trust Region Policy & 1. On-policy and actor-critic. \\
Optimisation (TRPO)~\cite{schulman_trust_2017} & 2. Uses a stochastic policy and learns from sampled trajectories. \\
& 3. Constrains the policy update to a KL-divergence based trust region to ensure stable learning. \\
& 4. Uses conjugate gradient descent~\cite{shewchuk_introduction_1994} to solve the constrained optimisation problem. \\ \midrule
Proximal Policy & 1. On-policy and actor-critic. \\
Optimisation (PPO)~\cite{schulman_proximal_2017} & 2. Uses a stochastic policy and learns from sampled trajectories. \\
& 3. Constrains the policy update using a clipped surrogate objective function. \\
& 4. Less computationally expensive than TRPO. \\ \midrule
Soft-Actor Critic (SAC) & 1. Off-policy and actor-critic. \\
\cite{haarnoja_soft_2018} & 2. Uses a stochastic policy and learns from an experience replay buffer. \\
& 3. Optimises an entropy based objective function, balancing reward and entropy. \\
& 4. Updates the policy, two Q-value networks, and an entropy parameter. \\ \midrule
Truncated Quantile & 1. Off-policy and actor-critic. \\
Critics (TQC)~\cite{kuznetsov_controlling_2020} & 2. Uses quantile regression to estimate the distribution of returns instead of the mean. \\
& 3.~Truncates quantiles in the critic's loss function to reduce impact of overestimations. \\
& 4. Employs a quantile Huber loss function to learn the quantile values. \\ 
\bottomrule
\end{tabular}
\end{table}

\clearpage

\subsection{RL Environments}

\begin{figure}[!h]
\begin{subfigure}{0.35\textwidth}
\centering
\includegraphics[height=5cm]{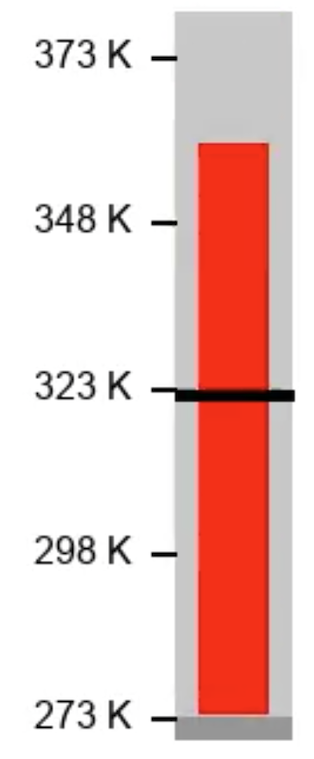}
\subcaption{Thermometer}
\end{subfigure}
\begin{subfigure}{0.65\textwidth}
\centering
\includegraphics[height=5cm]{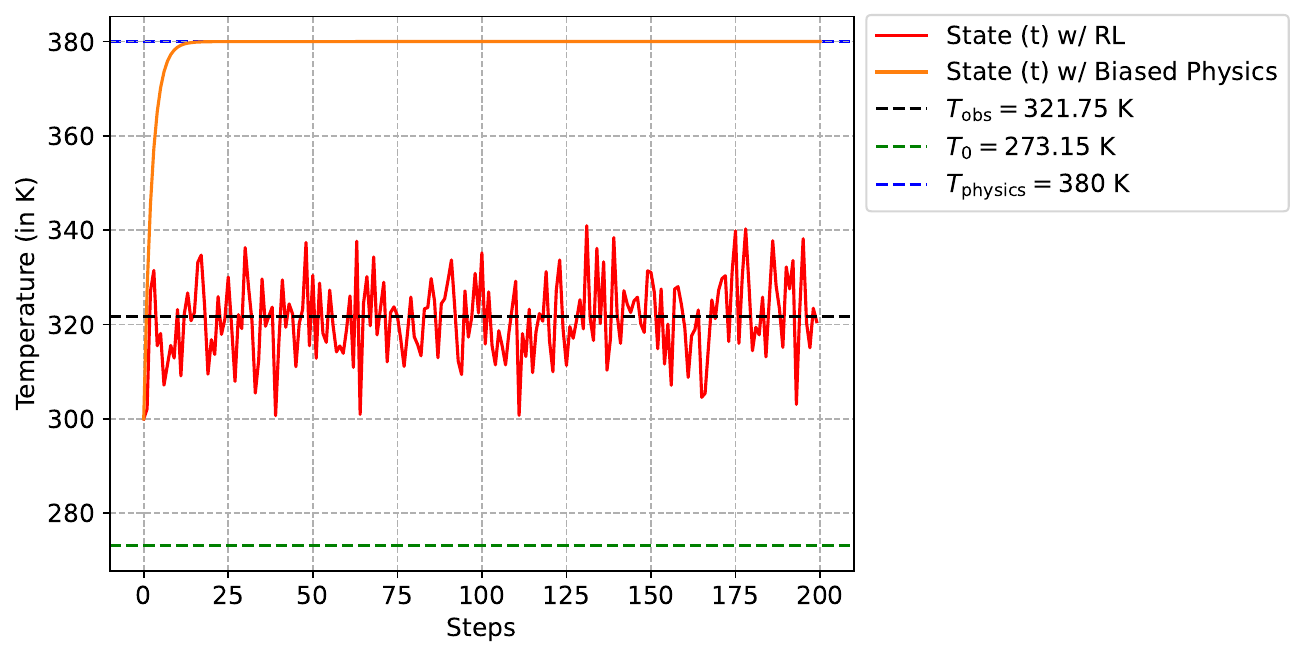}
\subcaption{Temperature evolution}
\end{subfigure}
\caption{(a) Thermometer (with 25 K marking intervals) visualising current temperature (indicated by red). Black line in the middle indicates the desired observed temperature of 321.75 K. (b) Line plot describing the state (temperature) evolution of both the RL agent and the biased physics model over 200 steps in one episode.}
\label{fig:scbc_thermometer}
\end{figure}

\begin{figure}[!h]
\centering
\begin{subfigure}[b]{0.48\textwidth}
\centering
\includegraphics[height=7.5cm]{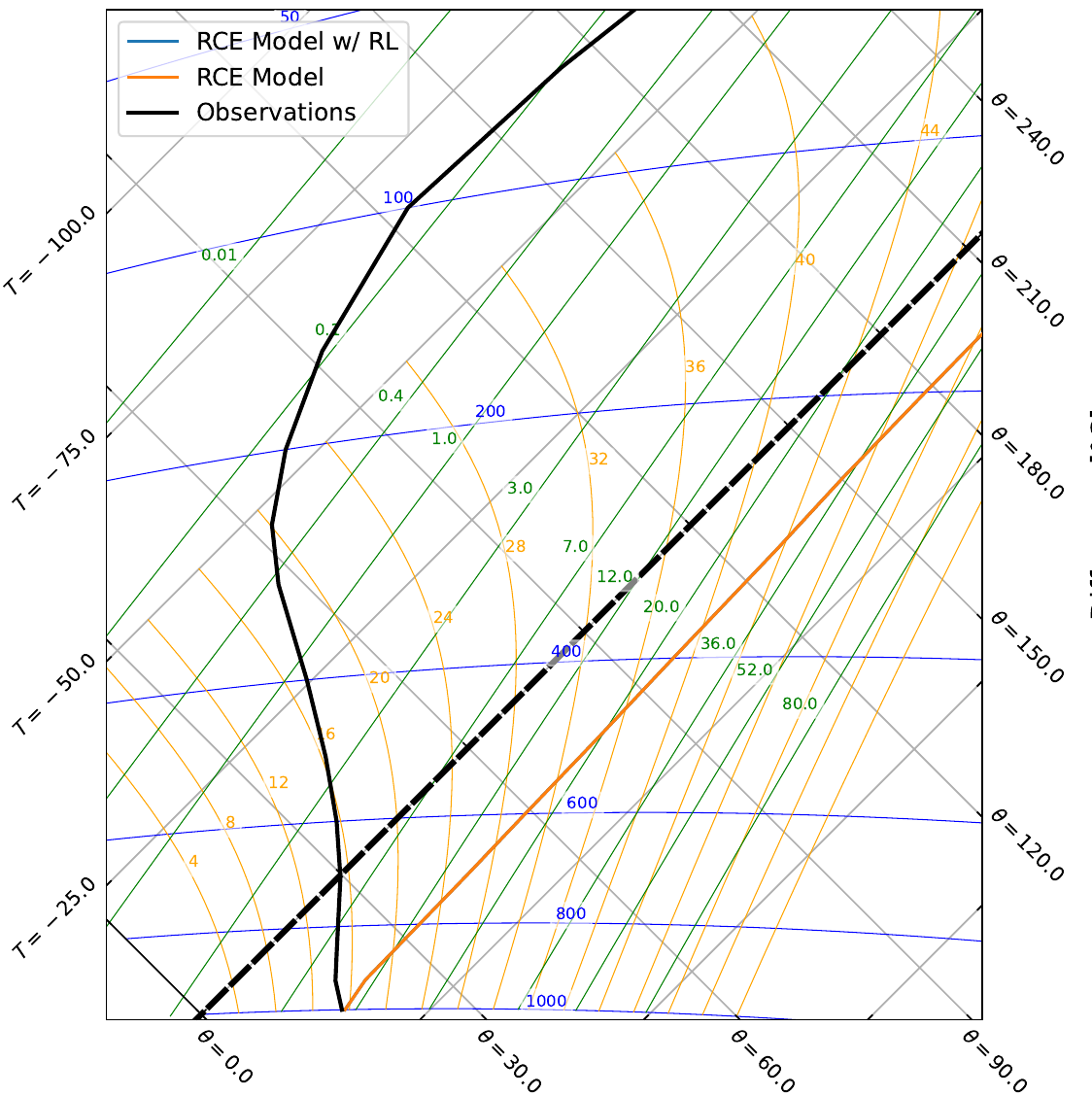}
\subcaption{Isothermal equilibrium at \texttt{step=0}}
\end{subfigure}
\begin{subfigure}[b]{0.48\textwidth}
\centering
\includegraphics[height=7.5cm]{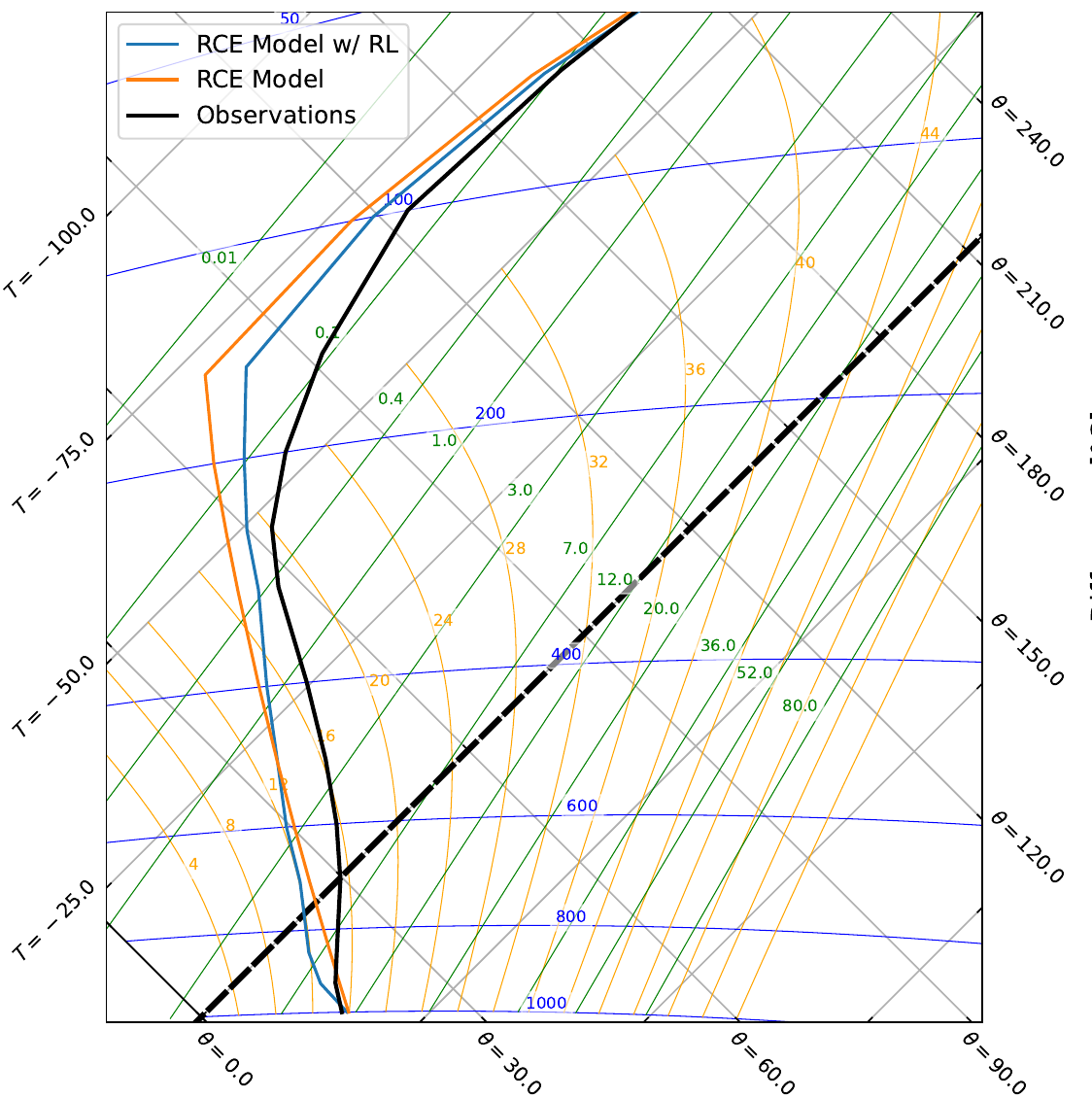}
\subcaption{Intermediate state at \texttt{step=200}}
\end{subfigure}
\caption{Tephigrams displaying three different temperature profiles: RCE Model with RL (blue), RCE Model (orange), and Observations (black).}
\label{fig:rcem_thermplots}
\end{figure}

Thermodynamic lines included in the Tephigrams (Fig.~\ref{fig:rcem_thermplots}) are the following:
\begin{enumerate}
\itemsep0em 
\item \textbf{Isobars}: Lines of constant pressure (in hPa). Displayed as dark blue curved left-right lines.
\item \textbf{Isotherms}: Lines of constant temperature (in $^{\circ}$C). Displayed as grey lines inclined at 45$^{\circ}$ (anti-clockwise).
\item \textbf{Dry adiabats}: Lines of constant potential temperature (in $^{\circ}$C). Displayed as grey lines inclined at 135$^{\circ}$ (anti-clockwise).
\item \textbf{Moist adiabats}: Lines of constant equivalent potential temperature (in $^{\circ}$C) for water saturated air parcels. Displayed as curved light orange lines from bottom to left.
\item \textbf{Saturated mixing ratio}: Lines of constant saturated mixing ratio (in g/kg). Displayed as green lines from bottom to top-right.  
\item \textbf{Isotherm at T $=$ 0$^{\circ}$C}: Line of constant temperature T $=$ 0$^{\circ}$C. Displayed as a thick dashed black line inclined at 45$^{\circ}$ (anti-clockwise).
\end{enumerate}

\begin{table}[!h]
\centering
\small
\ttfamily
\caption{Tabular representation of different RL hyperparameters}
\vspace{0.25em}
\label{tbl:tune_parameters}
\begin{tabular}{lp{10cm}l}
\toprule
\textbf{\textrm{Algorithm}} & \textbf{\textrm{Parameter Names}} & \textbf{\textrm{Count}}\\
\midrule
\textrm{REINFORCE} & learning\_rate, actor\_critic\_layer\_size & 2 \\ \midrule
\textrm{DDPG} & learning\_rate, tau, batch\_size, exploration\_noise, policy\_frequency, noise\_clip, actor\_critic\_layer\_size & 7 \\ \midrule
\textrm{DPG} & learning\_rate, exploration\_noise, policy\_frequency, actor\_critic\_layer\_size & 4 \\ \midrule
\textrm{TD3} & learning\_rate, tau, batch\_size, policy\_noise, exploration\_noise, policy\_frequency, noise\_clip, actor\_critic\_layer\_size & 8 \\ \midrule
\textrm{PPO} & learning\_rate, num\_minibatches, update\_epochs, clip\_coef, max\_grad\_norm, actor\_critic\_layer\_size & 6 \\ \midrule
\textrm{TRPO} & learning\_rate, num\_minibatches, update\_epochs, clip\_coef, max\_grad\_norm, actor\_critic\_layer\_size & 6 \\ \midrule
\textrm{SAC} & tau, batch\_size, policy\_lr, q\_lr, policy\_frequency, target\_network\_frequency, noise\_clip, alpha, actor\_critic\_layer\_size & 9 \\ \midrule
\textrm{TQC} & tau, batch\_size, n\_quantiles, n\_critics, actor\_adam\_lr, critic\_adam\_lr, alpha\_adam\_lr, policy\_frequency, target\_network\_frequency, actor\_critic\_layer\_size & 10 \\ 
\bottomrule
\end{tabular}
\end{table}

\subsection{Hyperparameter Tuning Setup}
\label{sec:app-hyperparameter_tuning}

\begin{figure}[!h]
\centering
\ifshowcontent{
\includegraphics[width=14cm]{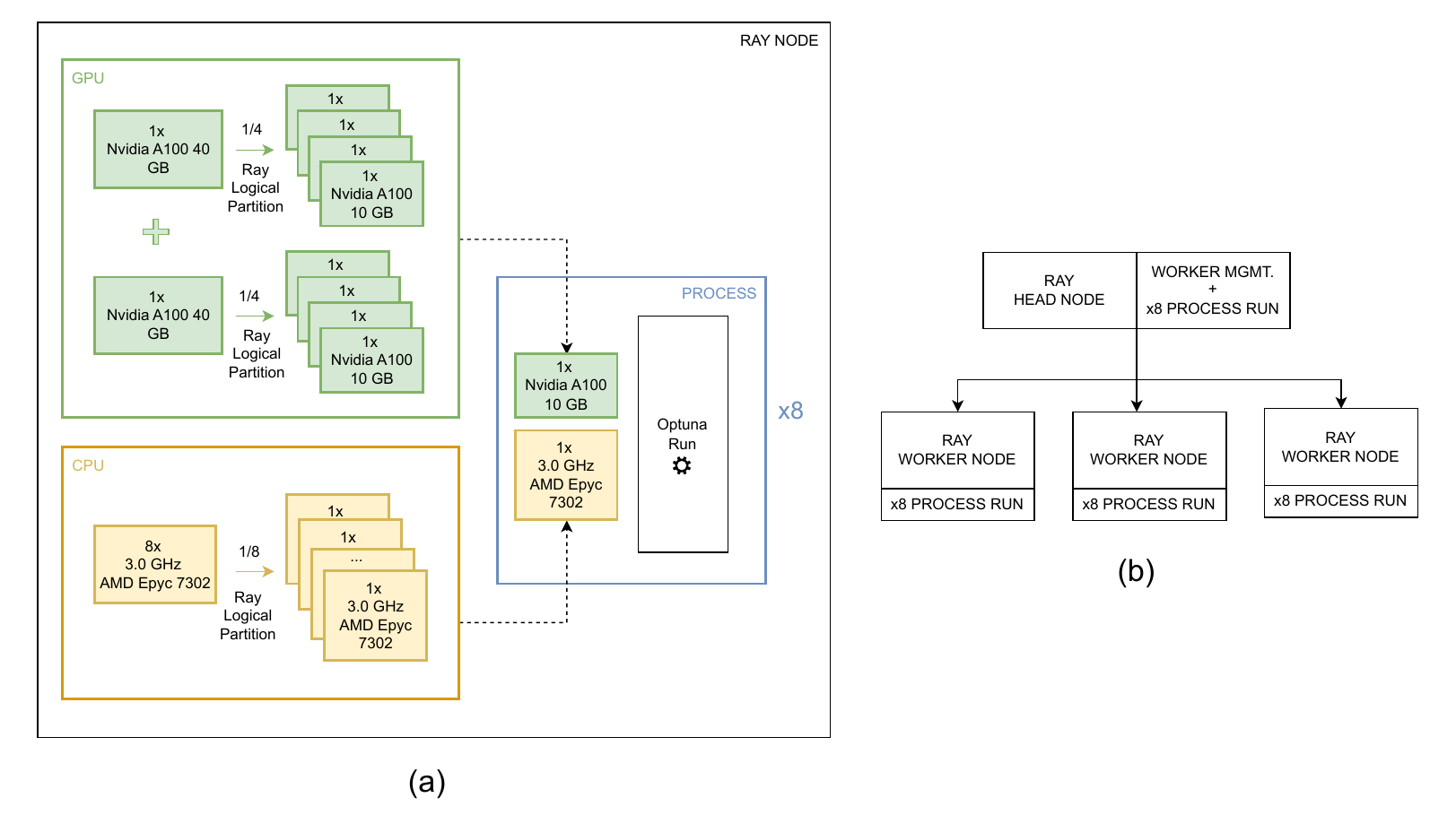}
}\fi
\ifnotshowcontent{
\includesvg[width=14cm]{param_ray_cluster.drawio.svg}
}\fi
\caption{Flow diagrams describing (a) the resource allocation in a Ray node and (b) the arrangement of Ray nodes in JASMIN}
\label{fig:param_ray_cluster}
\end{figure}

For tuning different hyperparameters (Table~\ref{tbl:tune_parameters}), we use Ray~\cite{moritz_ray_2018}, a distributed computing framework, interfacing with SLURM~\cite{goos_slurm_2003} to manage resources across one head node and three worker nodes. Each node is equipped with 8x 3.0 GHz AMD Epyc 7302 CPUs and 2x 40 GB Nvidia A100 GPUs. We partition each GPU into four logical units using Ray's resource management, enhancing parallel computation capabilities. We utilise Optuna~\cite{akiba_optuna_2019}, an automatic hyperparameter optimisation framework, for advanced parameter sampling and pruning of underperforming trials. This setup (Fig.~\ref{fig:param_ray_cluster}) allows us to initiate 32 concurrent tuning experiments for each RL algorithm, maximising the use of JASMIN's~\cite{lawrence_jasmin_2012} allocated resources per user. Such a high-performance parallel computing configuration made of Ray, SLURM, and Optuna on JASMIN's infrastructure enables rapid experimentation and scaling, crucial for swift testing and analysis within a constrained timeframe of six hours combined over all eight RL algorithms for each experiment.

\subsection{Evaluation}
\label{sec:app-evaluation}

\enlargethispage{3\baselineskip}

We evaluate RL environments using two parameter configurations: \texttt{optim-L} (optimised actor-critic layer size) and \texttt{homo-64L} (constant actor-critic layer size of 64). Each configuration has two runs: an ideal compute scenario (60k steps for \texttt{SimpleClimateBiasCorrectionEnv}, 10k for \texttt{RadiativeConvectiveModelEnv}) and a realistic compute scenario \cite{palmer_stochastic_2019}, outlining 16 experiments in total (Table~\ref{tbl:eval_experiments}). Our evaluation strategy focuses on three key aspects: a) $N_{\text{to\_threshold}}$: Steps to reach a specific episodic return threshold (Table~\ref{tbl:eval_thresholds}). b) $\sigma_{\text{after\_threshold}}^2$: Variance in episodic returns after reaching the threshold. c) $\Delta_{\text{from\_60k/10k}}$: Difference between threshold episodic return and final return in the ideal scenario. This approach is key in identifying RL algorithms combining sample efficiency (a, c) and stability (b) alongside addressing the lack of a standard train/test procedure, present in traditional ML.

\clearpage

\begin{table}[!h]
\centering
\caption{Experiment codes for each RL environment}
\vspace{0.25em}
\label{tbl:eval_experiments}
\ttfamily
\begin{tabular}{ll}
\toprule
\textbf{\textrm{Environment}} & \textbf{\textrm{Experiment ID}} \\ \midrule
\multirow{4}{*}{SimpleClimateBiasCorrection-v0} & v0-optim-L \\
 & v0-optim-L-60k \\
 & v0-homo-64L \\
 & v0-homo-64L-60k \\ \midrule
\multirow{4}{*}{SimpleClimateBiasCorrection-v1} & v1-optim-L \\
 & v1-optim-L-60k \\
 & v1-homo-64L \\
 & v1-homo-64L-60k \\ \midrule
\multirow{4}{*}{SimpleClimateBiasCorrection-v2} & v2-optim-L \\
 & v2-optim-L-60k \\
 & v2-homo-64L \\
 & v2-homo-64L-60k \\ \midrule
\multirow{4}{*}{RadiativeConvectiveModel-v0} & rce-v0-optim-L \\
 & rce-v0-optim-L-10k \\
 & rce-v0-homo-64L \\
 & rce-v0-homo-64L-10k \\ \bottomrule
\end{tabular}
\end{table}

\begin{table}[!h]
\centering
\caption{Empirically determined episodic return thresholds from the learning curves for each RL environment}
\vspace{0.25em}
\label{tbl:eval_thresholds}
\ttfamily
\begin{tabular}{lll}
\toprule
\textbf{\textrm{Environment}} & \textbf{\textrm{Threshold}} & \textbf{\textrm{Error (in K) per Episodic Step\footnotemark[1]}} \\ \midrule
{SimpleClimateBiasCorrection-v0} & -0.25 & $\pm$ 0.035 \\ 
{SimpleClimateBiasCorrection-v1} & -2.718 & $\pm$ 0.116 \\ 
{SimpleClimateBiasCorrection-v2} & -1 * (160\footnotemark[2] + 2.718) & $\pm$ 0.116 \\ 
{RadiativeConvectiveModel-v0} & -43900 & $\pm$ 9.37 (0.55\footnotemark[3]) \\
\bottomrule
\end{tabular}
\end{table}

\clearpage

\section{Additional Results}
\subsection{Result Generation Process}

\begin{figure}[!h]
\centering
\ifshowcontent{
\includegraphics[width=14cm]{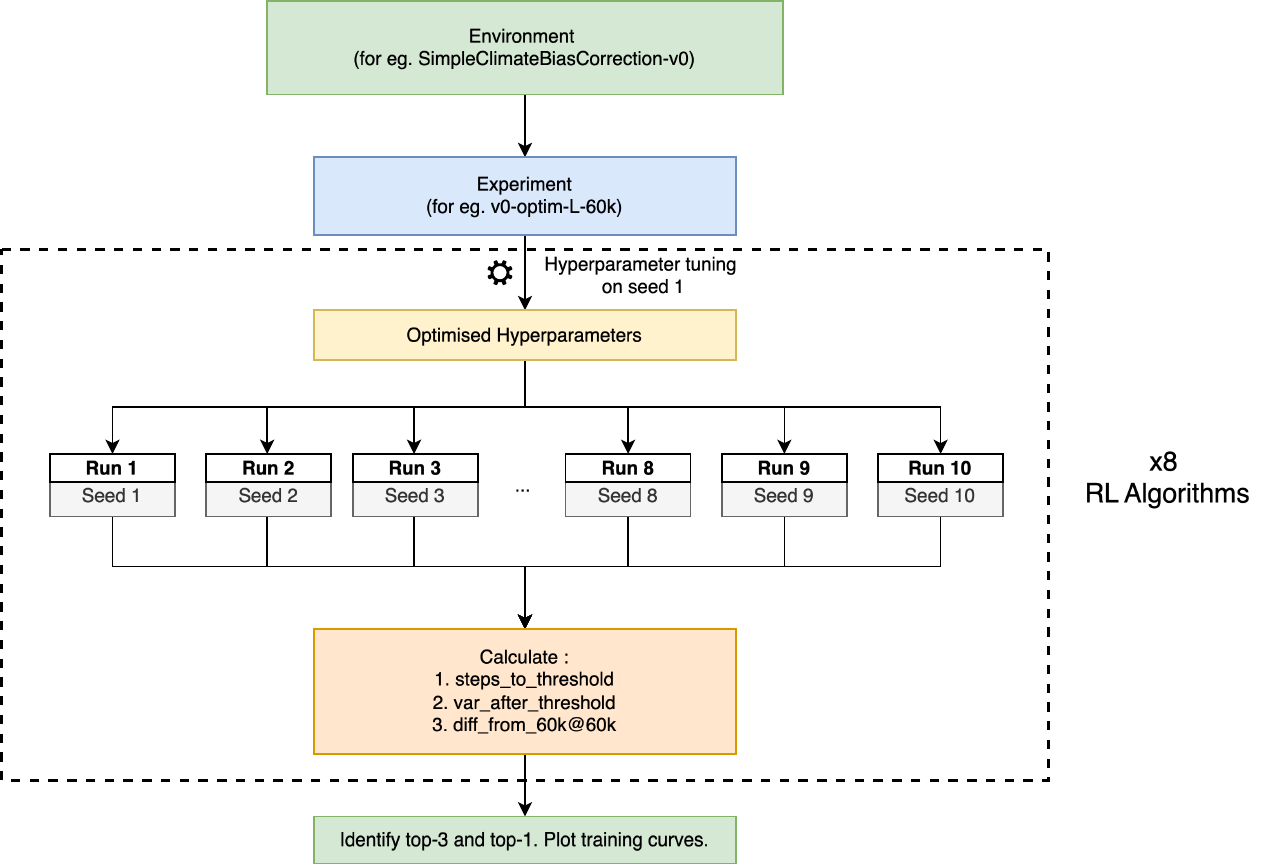}
}\fi
\ifnotshowcontent{
\includesvg[width=15cm]{experiment-process.drawio.svg}
}\fi
\caption{Flow diagram describing the end-to-end result generation process for each experiment}
\label{fig:results_overview}
\end{figure}

Our process to generate results (Fig.\ref{fig:results_overview}), begins with determining optimal hyperparameters using the Ray-powered parallel computing strategy (Section\ref{sec:app-hyperparameter_tuning}). After identifying these parameters with seed 1, we conduct 10 runs using random seeds (1-10) to capture variance in episodic returns, accounting for stochasticity in neural network weights at initialisation. We then calculate scores for three components (Section \ref{sec:app-evaluation}) across these 10 runs for each RL algorithm (Table~\ref{tbl:rl_four_point_summaries}). Based on the scores obtained, we report the top-3 and top-1 algorithms. This comprehensive approach ensures robust evaluation of RL algorithm performance across multiple experiments, considering both optimal parameters and performance variability.

\clearpage

\subsection{Top-3 Algorithms}

\begin{table}[!h]
\centering
\caption{Top-3 algorithms in \texttt{SimpleClimateBiasCorrectionEnv}}
\vspace{0.25em}
\label{tbl:results_scbc}
\ttfamily
\begin{tabular}{lllll}
\toprule
\textbf{\textrm{Environment}} & \textbf{\textrm{Experiment ID}} & \textbf{\textrm{\#1}} & \textbf{\textrm{\#2}} & \textbf{\textrm{\#3}} \\ 
\midrule
\multirow{4}{*}{SimpleClimateBiasCorrection-v0} & v0-optim-L & \textrm{TD3} & \textrm{TQC} & \textrm{DPG} \\
 & v0-optim-L-60k & \textrm{DDPG} & \textrm{TD3} & \textrm{TQC} \\
 & v0-homo-64L & \textrm{DPG} & \textrm{DDPG} & \textrm{TQC} \\
 & v0-homo-64L-60k & \textrm{TQC} & \textrm{DDPG} & \textrm{DPG} \\ 
\midrule
\multirow{4}{*}{SimpleClimateBiasCorrection-v1} & v1-optim-L & \textrm{TQC} & \textrm{DDPG} & \textrm{TD3} \\
 & v1-optim-L-60k & \textrm{TD3} & \textrm{DDPG} & \textrm{TQC} \\
 & v1-homo-64L & \textrm{DDPG} & \textrm{TD3} & \textrm{TQC} \\
 & v1-homo-64L-60k & \textrm{DDPG} & \textrm{TD3} & \textrm{TQC} \\ 
\midrule
\multirow{4}{*}{SimpleClimateBiasCorrection-v2} & v2-optim-L & \textrm{TD3} & \textrm{DDPG} & \textrm{DDPG} \\
 & v2-optim-L-60k & \textrm{TD3} & \textrm{SAC} & \textrm{DDPG} \\
 & v2-homo-64L & \textrm{TD3} & \textrm{DDPG} & \textrm{TQC} \\
 & v2-homo-64L-60k & \textrm{TD3} & \textrm{SAC} & \textrm{DDPG} \\ 
\bottomrule
\end{tabular}
\end{table}

\begin{table}[!h]
\centering
\caption{Top-3 algorithms in \texttt{RadiativeConvectiveModelEnv}}
\vspace{0.25em}
\label{tbl:results_rcem}
\ttfamily
\begin{tabular}{lllll}
\toprule
\textbf{\textrm{Environment}} & \textbf{\textrm{Experiment ID}} & \textbf{\textrm{\#1}} & \textbf{\textrm{\#2}} & \textbf{\textrm{\#3}} \\ 
\midrule
\multirow{4}{*}{RadiativeConvectiveModel-v0} & v0-optim-L & \textrm{DPG} & \textrm{DDPG} & \textrm{TQC} \\
 & v0-optim-L-10k & \textrm{DPG} & \textrm{PPO} & \textrm{TQC} \\
 & v0-homo-64L & \textrm{TRPO} & \textrm{PPO} & \textrm{DPG} \\
 & v0-homo-64L-10k & \textrm{TRPO} & \textrm{PPO} & \textrm{DPG} \\  
\bottomrule
\end{tabular}
\end{table}

\ifnotshowcontent{
\begin{table}[!h]
\centering
\caption{Frequency of RL algorithms within Top-3}
\label{tbl:results_topk}
\begin{subtable}{0.48\textwidth}
\centering
\caption{\texttt{SimpleClimateBiasCorrectionEnv}}
\label{tbl:results_scbc_topk}
\ttfamily
\begin{tabular}{ccc}
\toprule
\textbf{\textrm{Rank}} & \textbf{\textrm{Algorithm}} & \textbf{\textrm{Frequency}} \\ 
\midrule
1 & \textrm{DDPG} & 11 \\ 
2 & \textrm{TD3} & 10 \\ 
2 & \textrm{TQC} & 10 \\ 
4 & \textrm{DPG} & 3 \\ 
5 & \textrm{SAC} & 2 \\ 
\bottomrule
\end{tabular}
\end{subtable}
\begin{subtable}{0.48\textwidth}
\centering
\caption{\texttt{RadiativeConvectiveModelEnv}}
\label{tbl:results_rcem_topk}
\ttfamily
\begin{tabular}{ccc}
\toprule
\textbf{\textrm{Rank}} & \textbf{\textrm{Algorithm}} & \textbf{\textrm{Frequency}} \\ 
\midrule
1 & \textrm{DPG} & 4 \\ 
2 & \textrm{PPO} & 3 \\ 
3 & \textrm{TRPO} & 2 \\ 
3 & \textrm{TQC} & 2 \\
5 & \textrm{DDPG} & 1 \\
\bottomrule
\end{tabular}
\end{subtable}
\end{table}
}\fi

\begin{table}[!h]
\centering
\caption{Frequency of RL algorithms within Top-1}
\label{tbl:app-results_top1}
\begin{subtable}{0.48\textwidth}
\centering
\caption{\texttt{SimpleClimateBiasCorrectionEnv}}
\label{tbl:results_scbc_top1}
\ttfamily
\begin{tabular}{ccc}
\toprule
\textbf{\textrm{Rank}} & \textbf{\textrm{Algorithm}} & \textbf{\textrm{Frequency}} \\ 
\midrule
1 & \textrm{TD3} & 6 \\ 
2 & \textrm{DDPG} & 3 \\ 
3 & \textrm{TQC} & 2 \\ 
4 & \textrm{DPG} & 1 \\ 
\bottomrule
\end{tabular}
\end{subtable}
\begin{subtable}{0.48\textwidth}
\centering
\caption{\texttt{RadiativeConvectiveModelEnv}}
\label{tbl:results_rcem_top1}
\ttfamily
\begin{tabular}{ccc}
\toprule
\textbf{\textrm{Rank}} & \textbf{\textrm{Algorithm}} & \textbf{\textrm{Frequency}} \\ 
\midrule
1 & \textrm{DPG} & 2 \\ 
1 & \textrm{TRPO} & 2 \\ 
\bottomrule
\end{tabular}
\end{subtable}
\end{table}

\clearpage

\subsection{Training Curves w/ Error Bounds}

\begin{figure}[!h]
\centering
\begin{subfigure}[b]{0.48\textwidth}
\centering
\includegraphics[height=4.75cm]{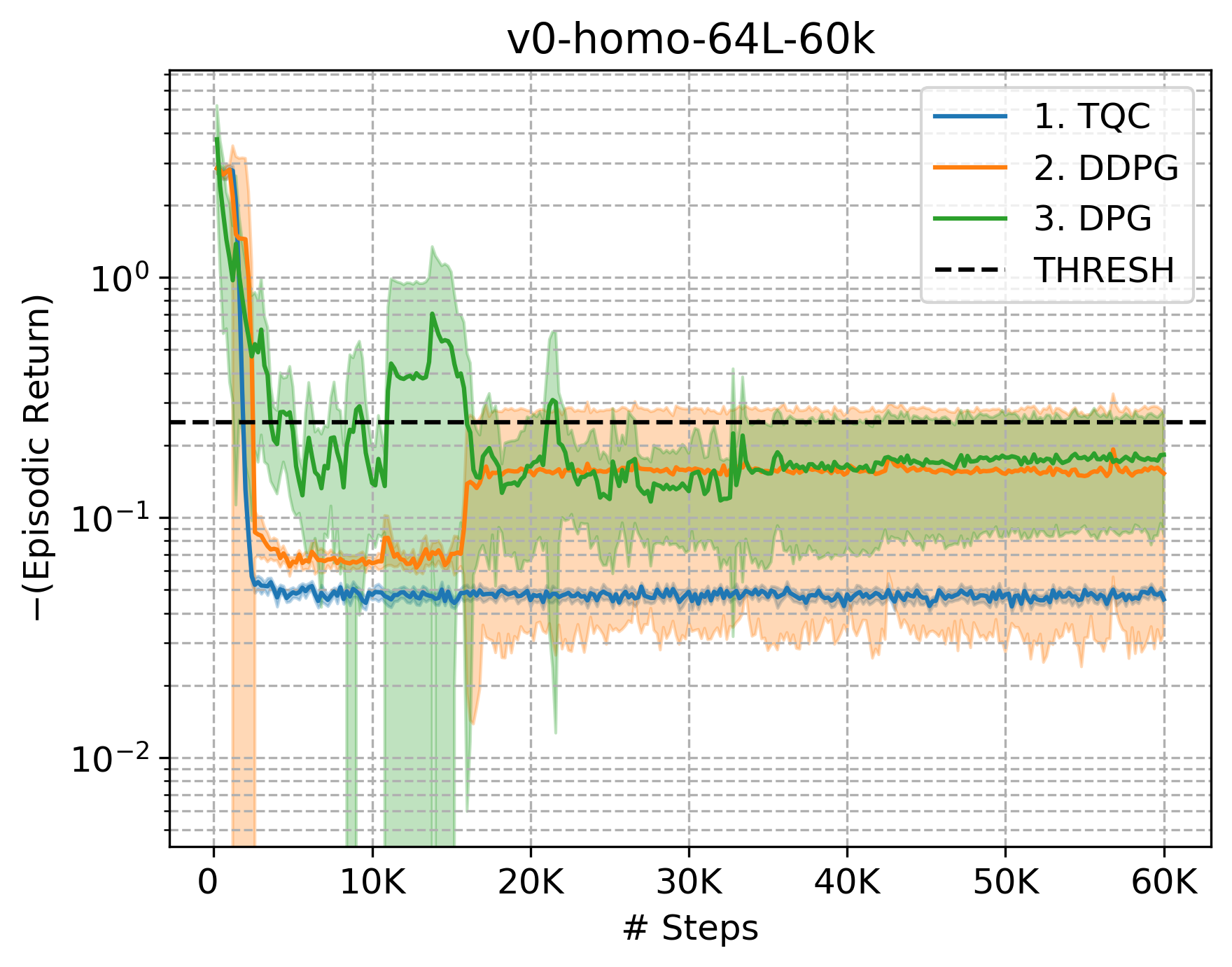}
\end{subfigure}
\begin{subfigure}[b]{0.48\textwidth}
\centering
\includegraphics[height=4.755cm]{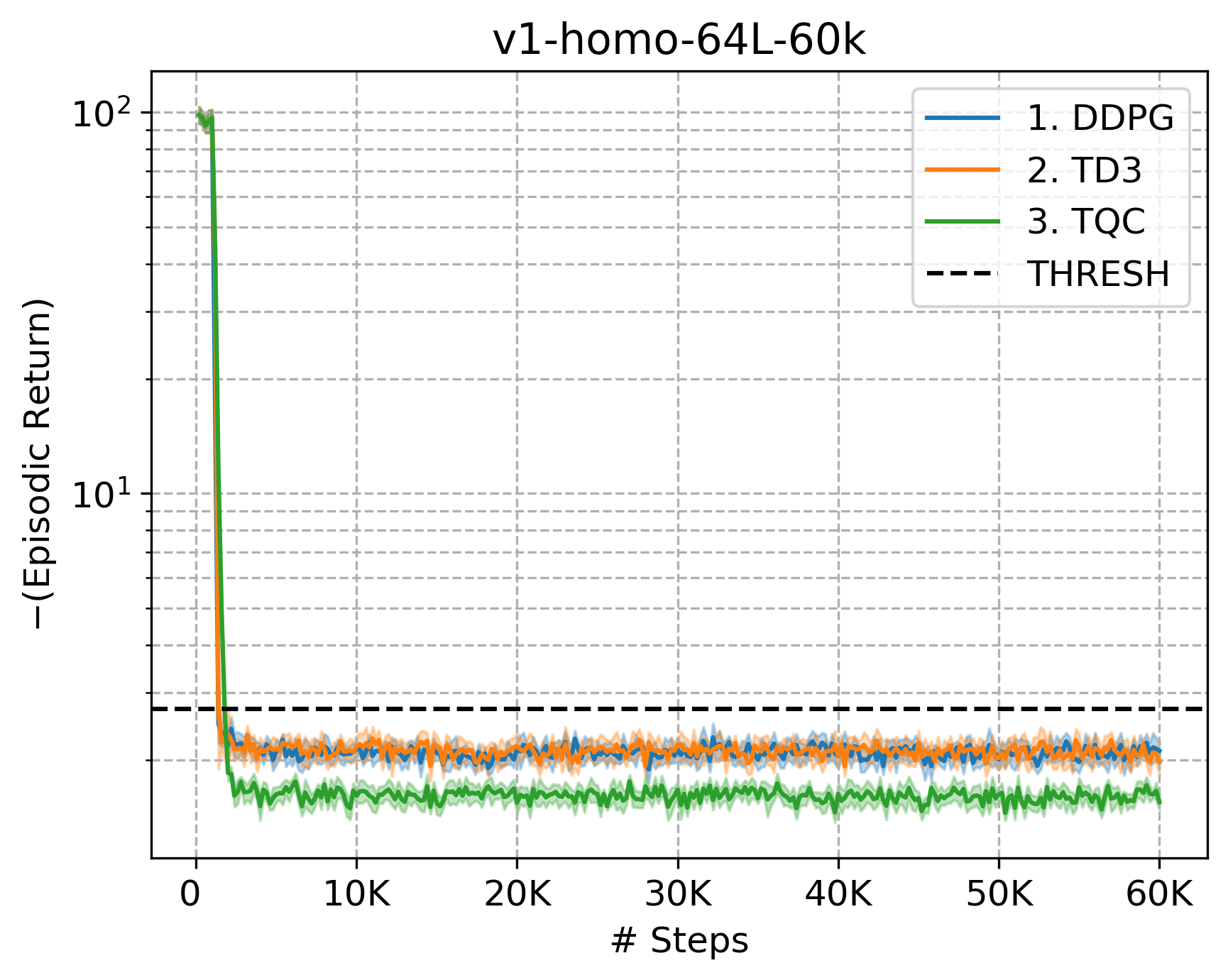}
\end{subfigure}
\begin{subfigure}[b]{0.48\textwidth}
\centering
\includegraphics[height=4.75cm]{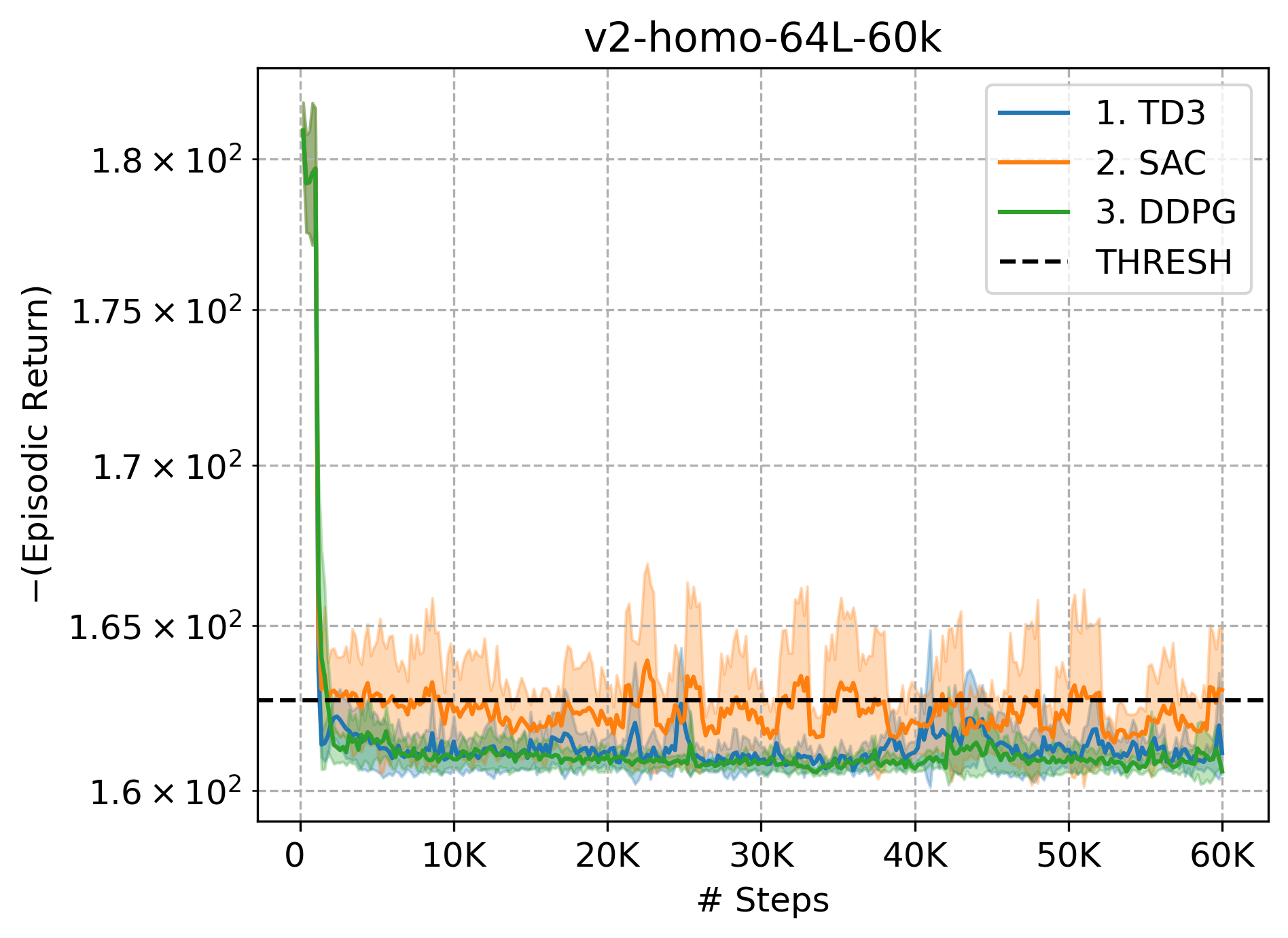}
\end{subfigure}
\caption{Top-3 RL algorithms with 95\% confidence intervals around the mean line for each \texttt{homo-64L-60k} experiment over 60k steps in \texttt{SimpleClimateBias\allowbreak CorrectionEnv} (1 episode = 200 steps).} 
\label{fig:results_scbc_graphs}
\end{figure}

\begin{figure}[!h]
\centering
\includegraphics[height=9cm]{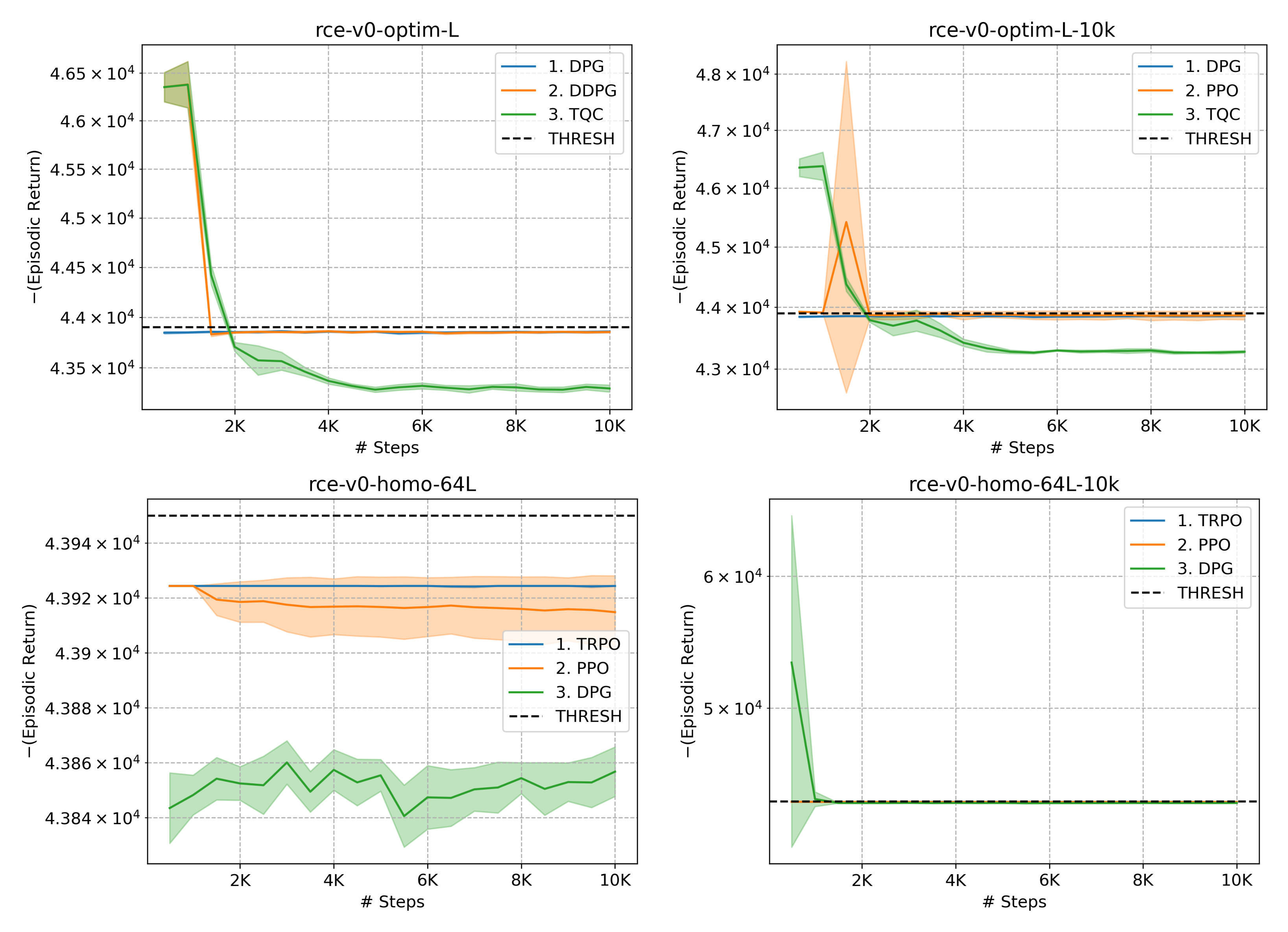}
\caption{Top-3 RL algorithms with 95\% confidence intervals around the mean line for each \texttt{RadiativeConvectiveModelEnv} experiment over 10k steps (1 episode = 500 steps). }
\label{fig:results_rcem_graphs}
\end{figure}

\begin{figure}[!h]
\centering
\includegraphics[width=\linewidth]{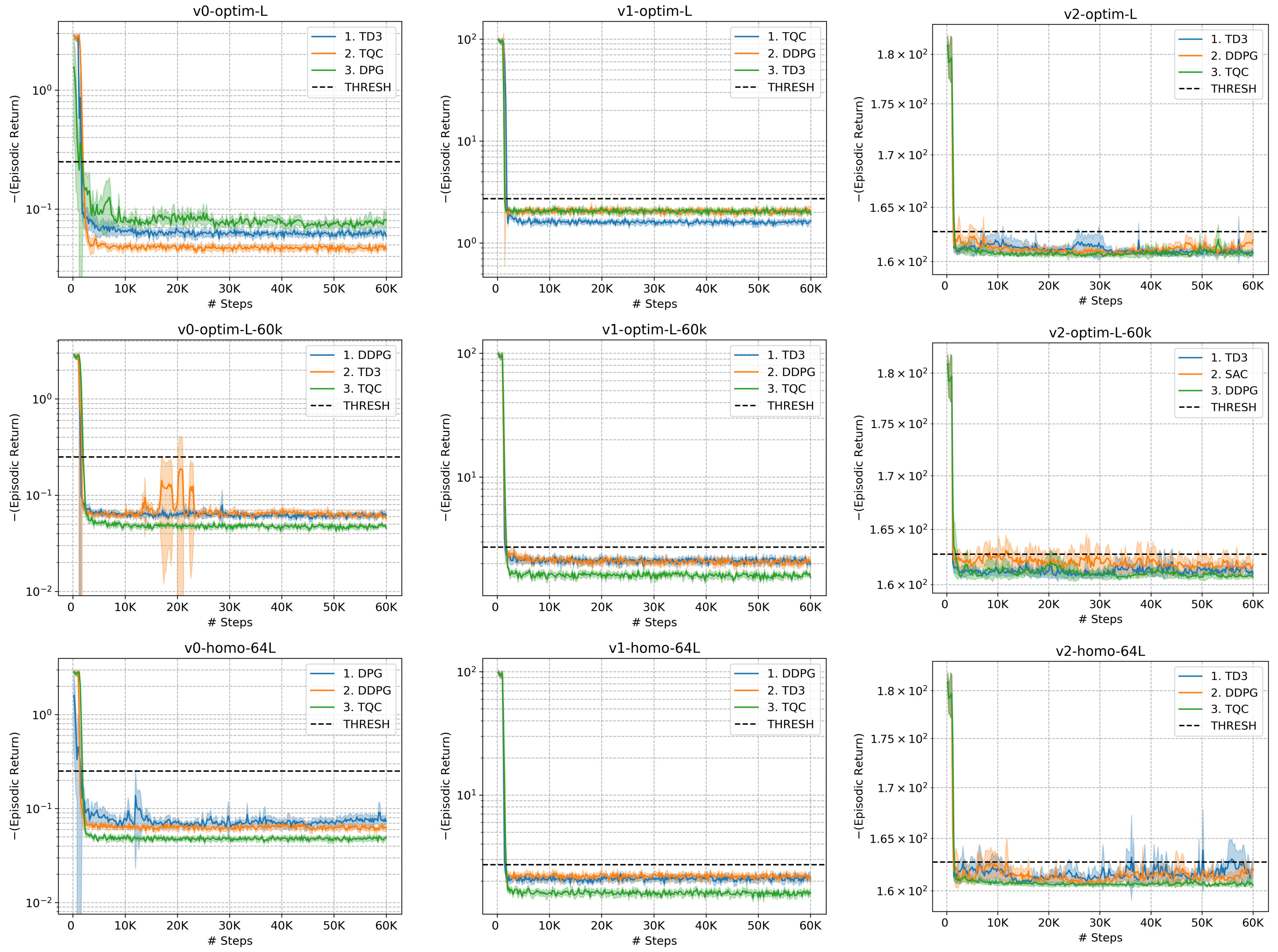}
\caption{Top-3 RL algorithms with 95\% confidence intervals around the mean line for each \texttt{SimpleClimateBiasCorrectionEnv} experiment (except \texttt{homo-64L-60k}) over 60k steps (1 episode = 200 steps).}
\label{fig:app-results_scbc_graphs}
\end{figure}

\subsection{Policy Analysis}

Analysis of policy decisions in the \texttt{SimpleClimateBiasCorrectionEnv} reveals a clear inverse correlation between the current state at timestep $t$ and the subsequent action at timestep $t+1$. This pattern aligns with the intuitive expectation that when temperature exceeds the normalised observed temperature threshold, the heating rate should be adjusted negatively (increased cooling) to reduce temperature, and vice-versa when temperature falls below the threshold. The consistency of this behaviour across all three versions of the environment over the experiments performed, suggests that the RL agents managed to learn a robust and physically meaningful policy for temperature regulation. These findings further support the potential of RL in addressing climate model bias correction and also generate interpretable policy decisions in climate-related applications.

\begin{figure}[!h]
\centering
\begin{subfigure}[b]{0.32\textwidth}
\centering
\includegraphics[height=3.45cm]{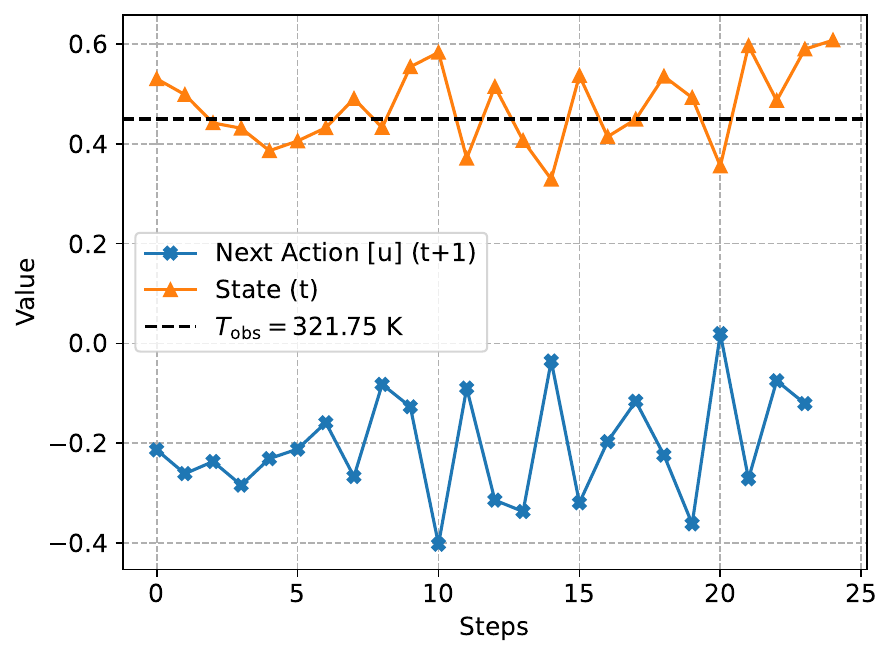}
\subcaption{\texttt{v0-optim-L-60k}\\DDPG}
\end{subfigure}
\begin{subfigure}[b]{0.32\textwidth}
\centering
\includegraphics[height=3.45cm]{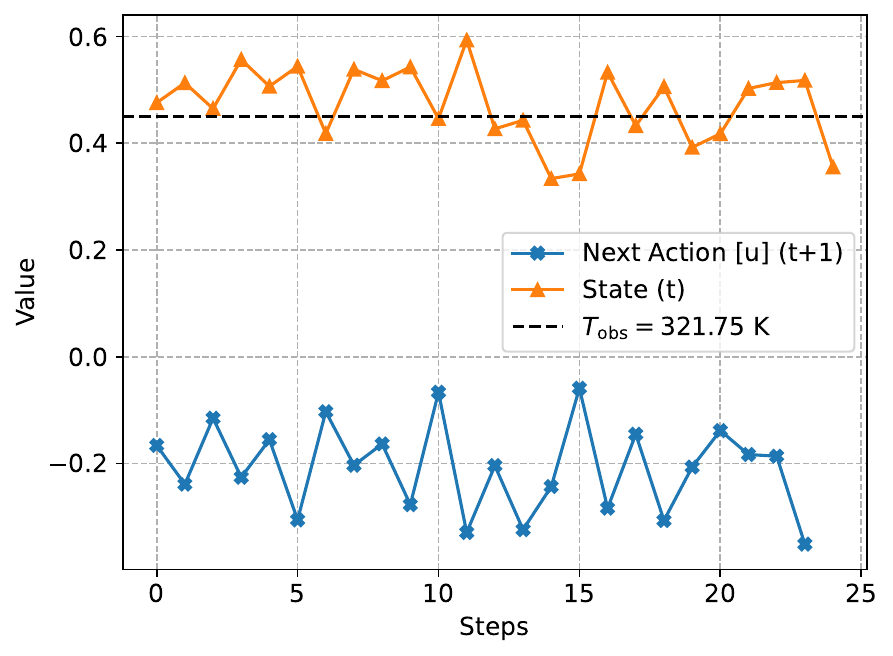}
\subcaption{\texttt{v1-optim-L-60k}\\TD3}
\end{subfigure}
\begin{subfigure}[b]{0.32\textwidth}
\centering
\includegraphics[height=3.45cm]{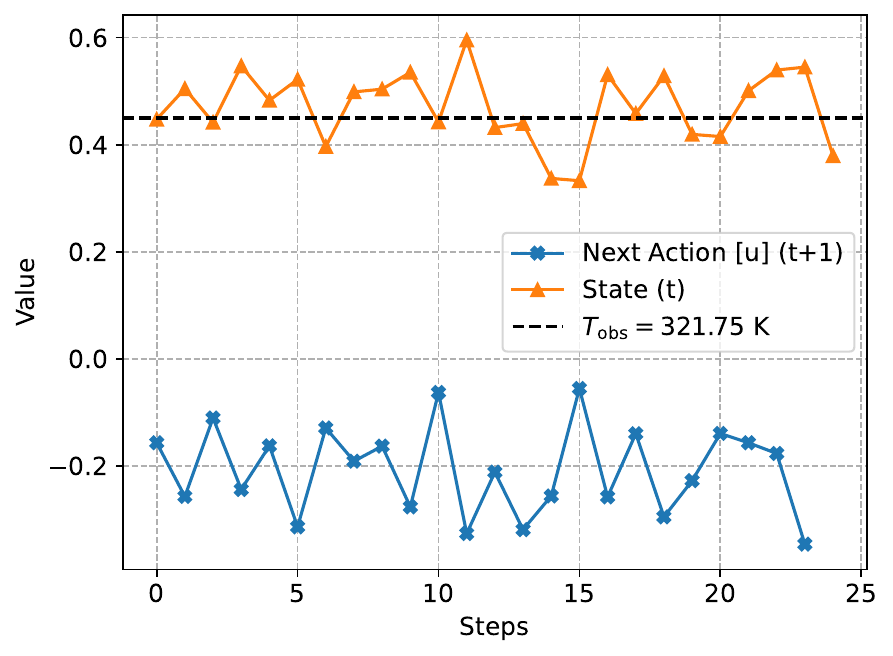}
\subcaption{\texttt{v2-optim-L-60k}\\TD3}
\end{subfigure}
\caption{Last 25 states (at time $t$) and their resulting RL agent actions (at time $t+1$) for \texttt{SimpleClimateBiasCorrection-v0/1/2}}
\label{fig:scbc_actions}
\end{figure}

\section{Algorithm Pseudocode}

\subsection{REINFORCE}

\begin{algorithm}[!h]
\caption{REINFORCE}
\label{alg:REINFORCE}
\begin{algorithmic}[1]
\State \textbf{Input:} Gym environment, Number of episodes $M$, Steps per episode $N$, Learning rate $\alpha$, Discount factor $\gamma$

\State \textbf{Initialise:} Policy network parameters $\theta$, Actor network $\pi_{\theta}$, Learning rate $\alpha$

\State \textbf{Pre-Setup:} Configure seed and environment variables, prepare environment and logging \\

\For{$episode = 1$ \textbf{to} $M$}
    \State Initialise episode buffer $B \leftarrow \emptyset$
    \State Observe initial state $s_0$
    \For{$t = 0$ \textbf{to} $N-1$}
        \State Select action $a_t \sim \pi_{\theta}(s_t)$
        \State Execute action $a_t$ and observe reward $r_t$ and new state $s_{t+1}$
        \State Store transition $(s_t, a_t, r_t)$ in $B$
        \State $s_t \leftarrow s_{t+1}$
    \EndFor
    \State $G \leftarrow 0$, $\mathcal{L}(\theta) \leftarrow 0$
    \For{$t$ \textbf{in} $B$ \textbf{reversed}}
    \State $G \leftarrow r_t + \gamma G$
    \State $\mathcal{L}(\theta) \leftarrow \mathcal{L}(\theta) - \nabla_{\theta} G \log \pi_{\theta}(a_t|s_t)$
    \EndFor
    \State Update policy parameters $\theta$ using accumulated gradients:
    \State \[
    \theta \leftarrow \theta +  \eta \frac{1}{|B|} \mathcal{L}(\theta)
    \]
\EndFor
\end{algorithmic}
\end{algorithm}

\clearpage

\subsection{Deterministic Policy Gradient (DPG)}

\begin{algorithm}[!h]
\caption{Deterministic Policy Gradient (DPG)}
\label{alg:DPG}
\begin{algorithmic}[1]
\State \textbf{Input:} Gym environment, Total timesteps $T$, Discount factor $\gamma$, Learning rate for policy $\eta_{\pi}$, Learning rate for Q-network $\eta_Q$, Batch size $B$, Exploration noise $\sigma$

\State \textbf{Initialise:} Policy network parameters $\theta$, Q-function network parameters $\phi$

\State \textbf{Pre-Setup:} Configure seed and environment variables, prepare environment and logging \\

\For{$t = 1$ \textbf{to} $T$}
    \State Observe state $s$ and select action $a = \pi_{\theta}(s) + \epsilon$, where $\epsilon \sim \mathcal{N}(0, \sigma)$
    \State Execute action $a$ and observe next state $s'$, reward $r$, and termination signal $d$
    \State Calculate Q-values:
    \State \[
    y(r, s', d) = r + \gamma (1 - d) Q_{\phi}(s', \pi_{\theta}(s'))
    \]
    \State Update Q-function by minimising the loss:
    \State \[
    \phi \gets \phi - \eta_Q \nabla_{\phi} \left(Q_{\phi}(s, a) - y(r, s', d)\right)^2
    \]
    \State Update policy by one step of gradient ascent:
    \State \[
    \theta \gets \theta + \eta_{\pi} \nabla_{\theta} Q_{\phi}(s, \pi_{\theta}(s))
    \]
\EndFor

\end{algorithmic}
\end{algorithm}

\clearpage

\subsection{Deep Deterministic Policy Gradient (DDPG)}

\begin{algorithm}[!h]
\caption{Deep Deterministic Policy Gradient (DDPG)}
\label{alg:DDPG}
\begin{algorithmic}[1]

\State \textbf{Input:} Gym environment, Total timesteps $T$, Replay buffer size $N$, Discount factor $\gamma$, Target smoothing coefficient $\tau$, Batch size $B$, Learning rate $\eta$, Exploration noise $\sigma$

\State \textbf{Initialise:} Policy network parameters $\theta$, Q-function network parameters $\phi$, target network parameters $\theta_{\text{targ}}$, $\phi_{\text{targ}}$, empty replay buffer $\mathcal{D}$

\State \textbf{Pre-Setup:} Configure seed and environment variables, prepare environment and logging \\

\For{$t = 1$ \textbf{to} $T$}
    \State Observe state $s$ and select action $a = \pi_{\theta}(s)$ \State Add exploration noise $a \gets a + \epsilon$, where $\epsilon \sim \mathcal{N}(0, \sigma)$ if required
    \State Execute action $a$ and observe next state $s'$, reward $r$, and termination signal $d$
    \State Store transition $(s, a, r, s', d)$ in $\mathcal{D}$
    \If{$t \geq \text{learning\_starts}$}
        \State Sample a minibatch of $B$ transitions $(s, a, r, s', d)$ from $\mathcal{D}$
        \State Compute target for Q-function update:
        \State \[
        y(r, s', d) = r + \gamma (1 - d) Q_{\phi_{\text{targ}}}(s', \pi_{\theta_{\text{targ}}}(s'))
        \]
        \State Update Q-function by minimising the loss:
        \State \[
        \phi \gets \phi - \eta \nabla_{\phi} \frac{1}{|B|} \sum_{(s,a,r,s',d) \in B} \left( Q_{\phi}(s, a) - y(r, s', d) \right)^2
        \]
        \State Update policy by one step of gradient ascent:
        \State \[
        \theta \gets \theta + \eta \nabla_{\theta} \frac{1}{|B|} \sum_{s \in B} Q_{\phi}(s, \pi_{\theta}(s))
        \]
        \State Soft-update target networks:
        \State \[
        \theta_{\text{targ}} \gets \tau \theta + (1 - \tau) \theta_{\text{targ}},  \quad
        \phi_{\text{targ}} \gets \tau \phi + (1 - \tau) \phi_{\text{targ}}
        \]
    \EndIf
\EndFor

\end{algorithmic}
\end{algorithm}

\clearpage

\subsection{Twin Delayed DDPG (TD3)}

\begin{algorithm}[!h]
\caption{Twin Delayed DDPG (TD3)}
\label{alg:TD3}
\begin{algorithmic}[1]
\State \textbf{Input:} Gym environment, Total timesteps $T$, Learning rate $\eta$, Replay buffer size $N$, Discount factor $\gamma$, Target smoothing coefficient $\tau$, Batch size $B$, Policy noise $\sigma_{\pi}$, Noise clip $\sigma_{\text{clip}}$, Exploration noise $\sigma_{\text{exploration}}$, Policy update frequency $f_{\pi}$

\State \textbf{Initialise:} Actor network $\theta$, Critic networks $\phi_1$, $\phi_2$, Target networks $\theta_{targ}$, $\phi_{targ, 1}$, $\phi_{targ, 2}$, Empty replay buffer $\mathcal{D}$

\State \textbf{Pre-Setup:} Configure seed and environment variables, prepare environment and logging \\

\For{$t = 1$ \textbf{to} $T$}
    \State Observe state $s$ and select action $a = \pi_{\theta}(s)$ \State Add exploration noise $a \gets a + \epsilon$, where $\epsilon \sim \mathcal{N}(0, \sigma_{\text{exploration}})$ if required
    \State Execute action $a$ and observe next state $s'$, reward $r$, and done signal $d$
    \State Store transition $(s, a, r, s', d)$ in $\mathcal{D}$
    \If{$t \geq \text{learning\_starts}$}
        \State Sample a minibatch of $B$ transitions $(s, a, r, s', d)$ from $\mathcal{D}$
        \State Compute target actions:
        \State \[
        a' \leftarrow \pi_{\theta_{targ}}(s') + \text{clip}(\mathcal{N}(0, \sigma_{\pi}), -\sigma_{\text{clip}}, \sigma_{\text{clip}})
        \]
        \State Compute target Q-values:
        \State \[
        y(r, s', d) \leftarrow r + \gamma (1 - d) \min_{i=1,2} Q_{\phi_{targ, i}}(s', a')
        \]
        \State Update critic networks by minimising the loss:
        \State \[
        \quad \phi_i \leftarrow \phi_i - \eta \nabla_{\phi_i} \frac{1}{|B|} \sum_{(s,a,r,s',d) \in B} \left( Q_{\phi_i}(s, a) - y(r, s', d) \right)^2, \text{for } i = 1, 2 
        \]
        \If{$t \mod f_{\pi} = 0$}
            \State Update actor network by policy gradient:
            \State \[
            \theta \leftarrow \theta + \eta \nabla_{\theta} \frac{1}{|B|} \sum_{s \in B} Q_{\phi_1}(s, \pi_\theta(s))
            \]
            \State Soft update target networks:
            \State \[
            \theta_{targ} \leftarrow \tau \theta + (1 - \tau) \theta_{targ}, \quad \phi_{targ, i} \leftarrow \tau \phi_i + (1 - \tau) \phi_{targ, i} \text{ for } i=1,2
            \]
        \EndIf
    \EndIf
\EndFor
\end{algorithmic}
\end{algorithm}

\clearpage

\subsection{Trust Region Policy Optimization (TRPO)}

\begin{algorithm}[!h]
\caption{Trust Region Policy Optimization (TRPO)}
\label{alg:TRPO}
\begin{algorithmic}[1]
\State \textbf{Input:} Gym environment, Total timesteps $T$, Mini-batch size $M$, Number of steps per episode $N$, Discount factor $\gamma$, GAE lambda $\lambda$, KL divergence limit $\delta$, Trust region update size $\beta$

\State \textbf{Initialise:} Policy parameters $\theta$, Value function parameters $\phi$

\State \textbf{Pre-Setup:} Configure seed and environment variables, prepare environment and logging \\

\For{$iteration = 1, 2, \dots, \frac{T}{N}$}
    \State Collect set of trajectories $\mathcal{D} = \{\tau_i\}$ by running policy $\pi_\theta$ in the environment
    \State Compute returns $\{{R_i}\}$ and advantage estimates $\{\hat{A_i}\}$ using GAE

    \For{$epoch = 1, 2, \dots, K$}
        \State Shuffle $D$ to create $M$ mini-batches
        \For{each mini-batch $t$}
            \State Update value function by minimising the MSE loss: 
            \State 
            \[
            L(\phi) = \frac{1}{2M}\sum_{t} \left(V_\phi(s_t) - \hat{R}_t\right)^2
            \]
            \State Compute the surrogate objective (policy loss):
            \[
            L^{\pi}(\theta) = \frac{1}{M} \sum_{t} \frac{\pi_\theta(a_t|s_t)}{\pi_{\theta_{old}}(a_t|s_t)} \hat{A}_t
            \]
            \State Compute policy gradient $\nabla_{\theta} L^{\pi}(\theta)$
            \State Apply conjugate gradient to estimate the natural policy gradient $\hat{g}$
            \[
            \hat{g} \approx (\nabla_{\theta}^2 KL(\pi_{\theta_{old}} \| \pi_\theta))^{-1} \nabla_{\theta} L^{\pi}(\theta)
            \]
            \State Compute step size $\alpha$ using line search:
            \[
            \alpha = \sqrt{\frac{2\delta}{\hat{g}^T H \hat{g}}}, \text{ where $H$ is the Hessian of $KL(\pi_{\theta_{old}} \| \pi_\theta)$}
            \]
            \State Update policy $\theta \leftarrow \theta + \alpha \hat{g}$ using an exponential increment strategy
        \EndFor
        \State \textbf{break} if $KL(\pi_{\theta_{old}} \| \pi_\theta) > \delta$
    \EndFor
\EndFor

\end{algorithmic}
\end{algorithm}

\clearpage

\subsection{Proximal Policy Optimization (PPO)}

\begin{algorithm}[!h]
\caption{Proximal Policy Optimization (PPO)}
\label{alg:PPO}
\begin{algorithmic}[1]
\State \textbf{Input:} Gym environment, Total timesteps $T$, Number of steps per episode $N$, Mini-batch size $M$, Update epochs $K$, Learning rate $\alpha$, Discount factor $\gamma$, GAE lambda $\lambda$, Clipping parameter $\epsilon$, VF coefficient $c_1$, Entropy coefficient $c_2$, KL divergence limit $\delta$

\State \textbf{Initialise:} Policy parameters $\theta$, Value function parameters $\phi$

\State \textbf{Pre-Setup:} Configure seed and environment variables, prepare environment and logging \\

\For{$iteration = 1, 2, \dots, \frac{T}{N}$}
    \State Collect set of trajectories $D = \{\tau_i\}$ by running policy $\pi_\theta$ in the environment
    \State Compute returns $\{{R_i}\}$ and advantage estimates $\{\hat{A_i}\}$ using GAE
    \For{$epoch = 1, 2, \dots, K$}
        \State Shuffle $D$ to create $M$ mini-batches
        \For{each mini-batch $t$}
            \State Compute ratio $r_t(\theta) = \frac{\pi_\theta(a_t|s_t)}{\pi_{\theta_{old}}(a_t|s_t)}$
            \State Compute clipped surrogate objective (policy loss):
            \[
            L^{CLIP}(\theta) = \hat{\mathbb{E}}_t \left[ \min(r_t(\theta) \hat{A}_t, \text{clip}(r_t(\theta), 1-\epsilon, 1+\epsilon) \hat{A}_t) \right]
            \]
            \State Compute value function loss:
            \[
            L^{VF}(\phi) = \left( V_\phi(s_t) - \hat{R}_t \right)^2
            \]
            \State Compute entropy: $S[\pi_\theta](s_t)$
            \State Compute total loss:
            \[
            L(\theta, \phi) = -L^{CLIP}(\theta) + c_1 L^{VF}(\phi) - c_2 S[\pi_\theta](s_t)
            \]
            \State Update $\theta$ and $\phi$ using stochastic gradient descent
        \EndFor
    \State \textbf{break} if $KL(\pi_{\theta_{old}} \| \pi_\theta) > \delta$
    \EndFor
\EndFor
\end{algorithmic}
\end{algorithm}

\clearpage

\subsection{Soft Actor-Critic (SAC)}

\begin{algorithm}[!h]
\caption{Soft Actor-Critic (SAC)}
\label{alg:SAC}
\begin{algorithmic}[1]

\State \textbf{Input:} Gym environment, Total timesteps $T$, Replay buffer size $N$, Discount factor $\gamma$, Target smoothing coefficient $\tau$, Batch size $B$, Learning rate for policy $\eta_{\pi}$, Learning rate for Q-network $\eta_Q$

\State \textbf{Initialise:} Policy network parameters $\theta$, Critic network parameters $\phi_1$, $\phi_2$, Target critic parameters $\phi_{\text{targ},1}$, $\phi_{\text{targ},2}$, Empty replay buffer $\mathcal{D}$, actor $\pi_\theta$, Entropy coefficient $\alpha$, Target entropy coefficient $\alpha_{\text{targ}}$

\State \textbf{Pre-Setup:} Configure seed and environment variables, prepare environment and logging \\

\For{$t = 1$ \textbf{to} $T$}
    \State Observe state $s$ and select action $a \sim \pi_\theta(s)$ with exploration strategy if required
    \State Execute action $a$ and observe next state $s'$, reward $r$, and termination signal $d$
    \State Store transition $(s, a, r, s', d)$ in $\mathcal{D}$
    \If{$t \geq \text{learning\_starts}$}
        \State Sample a minibatch of $B$ transitions $(s, a, r, s', d)$ from $\mathcal{D}$
        \State Compute targets for critic updates:
        \State \[
        y(r, s', d) = r + \gamma (1 - d) \left( \min_{i=1,2} Q_{\phi_{\text{targ},i}}(s', \tilde{a}') - \alpha \log \pi_\theta(\tilde{a}' | s') \right) \]
        \State where $\tilde{a}' \sim \pi_\theta(s')$
        \State Update Q-functions by one step of gradient descent:
        \State \[
        \phi_i \gets \phi_i - \eta_Q \nabla_{\phi_i} \frac{1}{|B|} \sum_{(s,a,r,s',d) \in B} \left( Q_{\phi_i}(s, a) - y(r, s', d) \right)^2 \text{ for } i=1,2
        \]
        \State Update policy by one step of gradient ascent:
        \State \[
        \theta \gets \theta + \eta_{\pi} \nabla_\theta \frac{1}{|B|} \sum_{s \in B} \left( \min_{i=1,2} Q_{\phi_i}(s, \pi_\theta(s)) - \alpha \log \pi_\theta(a|s) \right)
        \]
        \State Soft-update target networks:
        \State \[
        \phi_{\text{targ},i} \gets \tau \phi_i + (1 - \tau) \phi_{\text{targ},i} \text{ for } i = 1, 2
        \]
        \State Optionally adjust $\alpha$ based on entropy targets:
        \State \[
        \alpha \gets \alpha + \eta_Q \nabla_{\alpha} \frac{\alpha}{|B|}\sum_{s \in B} \left( \log \pi_\theta(a|s) + \alpha_{\text{targ}} \right)
        \]
    \EndIf
\EndFor

\end{algorithmic}
\end{algorithm}

\clearpage

\subsection{Truncated Quantile Critics (TQC)}

\enlargethispage{2\baselineskip}

\begin{algorithm}[!h]
\caption{Truncated Quantile Critics (TQC)}
\label{alg:TQC}
\begin{algorithmic}[1]
\State \textbf{Input:} Gym environment, Total timesteps $T$, Replay buffer size $N$, Discount factor $\gamma$, Smoothing coefficient $\tau$, Batch size $B$, Learning rate $\eta$, Number of quantiles $N_q$, Number of critics $N_c$, Drop quantiles $N_{\text{drop}}$, Entropy coefficient $\alpha$, Target entropy coefficient $\alpha_{\text{targ}}$

\State \textbf{Initialise:} Actor network $\theta$, Critic network parameters $\phi_1, \dots, \phi_{N_c}$, Target critic network parameters $\phi_{\text{targ},1}, \dots, \phi_{\text{targ},N_c}$, Replay buffer $\mathcal{D}$

\State \textbf{Pre-Setup:} Configure seed and environment variables, prepare environment and logging \\

\For{$t = 1$ \textbf{to} $T$}
    \State Select action $a \sim \pi_\theta(s)$ based on current policy and exploration strategy
    \State Execute action $a$ and observe next state $s'$, reward $r$, and done signal $d$
    \State Store transition tuple $(s, a, r, s', d)$ in $\mathcal{D}$
    \If{$t \geq \text{learning\_starts}$} 
        \For{$i = 1$ \textbf{to} $N_c$}
            \State Sample a minibatch of $B$ transitions $(s, a, r, s', d)$ from $\mathcal{D}$
            \State Compute target quantile values for critic $\phi_{target, i}$:
            \State \[
        y(r, s', d) = r + \gamma (1 - d) \left( Q_{\phi_{\text{targ},i}}(s', \tilde{a}', N_{\text{drop}}) - \alpha \log \pi_\theta(\tilde{a}' | s') \right) \]
        \State where $\tilde{a}' \sim \pi_\theta(s')$        
            \State Update critic $\phi_i$ by minimising the quantile Huber loss:
            \[
            \text{L}^{\phi_i} = \frac{1}{N_q} \sum_{k=1}^{N_q} \text{HuberLoss}(Q_{\phi_i}(s_j, a_j, \tau_k) - y_j) \]
            \State where $\tau_k$ are the quantile fractions
        \EndFor
        \State Update policy by one step of gradient ascent:
        \State \[
        \theta \gets \theta + \eta \nabla_\theta \frac{1}{|B|} \sum_{s \in B} \left( - \alpha \log \pi_\theta(a|s) + \frac{1}{N_c} \sum_{i=1}^{N_c} Q_{\phi_i}(s, \pi_\theta(s)) \right)
        \]
        \State Soft-update target networks:
        \State \[
        \phi_{\text{targ},i} \gets \tau \phi_i + (1 - \tau) \phi_{\text{targ},i} \text{ for } i = 1, 2, ..., N_c
        \]
        \State Optionally adjust $\alpha$ based on entropy targets:
        \State \[
        \alpha \gets \alpha + \eta \nabla_{\alpha} \frac{\alpha}{|B|}\sum_{s \in B} \left( \log \pi_\theta(a|s) + \alpha_{\text{targ}} \right)
        \]
    \EndIf
\EndFor

\end{algorithmic}
\end{algorithm}

\end{document}